\pdfoutput=1
\documentclass[11pt]{article}

\PassOptionsToPackage{table,dvipsnames}{xcolor}
\usepackage[final]{acl}

\usepackage{times}
\usepackage{latexsym}
\usepackage[T1]{fontenc}
\usepackage[utf8]{inputenc}
\usepackage{microtype}
\usepackage{inconsolata}

\usepackage{graphicx}
\usepackage{colortbl}
\usepackage{array}
\usepackage{booktabs}
\usepackage{multirow}
\usepackage{arydshln}
\usepackage{adjustbox}
\usepackage{subcaption}

\usepackage{algorithm}
\usepackage{algpseudocode}

\usepackage{amsmath}
\usepackage{amsfonts}
\usepackage{nicefrac}
\usepackage{wasysym}
\usepackage{fontawesome}

\usepackage{enumitem}
\usepackage{url}
\usepackage{lipsum}
\usepackage{natbib}

\usepackage{tocloft}
\usepackage{minitoc}

\usepackage[most]{tcolorbox}
\tcbset{breakable}  

\usepackage{hyperref}

\usepackage{setspace}

\usepackage{xspace}
\usepackage{listings}

\newcommand{\approach}{\textsc{Self-Foveate}\xspace}%
\newcommand{\sotallm}{GPT-4o\xspace}%

\newcommand{\checkmarkcolor}{
    \tikz[baseline=(checkIcon.base)]{
        \node[draw=gray!30,line width=0.5pt, fill=green!70!black,rounded corners,inner sep=1.5pt] (checkIcon) {\textcolor{white}{$\checkmark$}};
    }
}

\newcommand{\checkmarkerror}{
    \tikz[baseline=(checkIcon.base)]{
        \node[draw=gray!30,line width=0.5pt, fill=gray,rounded corners,inner sep=1.5pt] (checkIcon) {\textcolor{white}{$\checkmark$}};
    }
}

\newcommand{\xmarkcolor}{%
  \textcolor{red}{%
    \begin{tikzpicture}[scale=0.2]
      \draw [line width=1.5] (0,0) -- (1,1);
      \draw [line width=1.5] (1,0) -- (0,1);
    \end{tikzpicture}%
  }%
}

\title{\approach: Enhancing Diversity and Difficulty of Synthesized Instructions from Unsupervised Text via Multi-Level Foveation}

\author{Mingzhe Li,\ \ \ Xin Lu,\ \ \ Yanyan Zhao\thanks{\ \ \  Email corresponding.} \\
	Research Center for Social Computing and Interactive Robotics\\
	Harbin Institute of Technology, China\\
	\texttt{\{mzli, xlu, yyzhao\}@ir.hit.edu.cn}
}

\begin{document}
\maketitle

\begin{abstract}
Synthesizing high-quality instruction data from unsupervised text is a promising paradigm for training large language models (LLMs), yet automated methods for this task still exhibit significant limitations in the diversity and difficulty of synthesized instructions. To address these challenges, we propose \approach, an LLM-driven method for instruction synthesis. Inspired by hierarchical human visual perception, \approach introduces a ``Micro-Scatter-Macro'' multi-level foveation methodology that guides the extraction of textual information at three complementary granularities, from fine-grained details through cross-region connections to holistic patterns, thereby enhancing both the diversity and difficulty of synthesized instructions. Furthermore, a re-synthesis module is incorporated to improve the fidelity of instructions to source text and their overall quality. Comprehensive experiments across multiple unsupervised corpora and diverse model architectures demonstrate that \approach consistently outperforms existing methods. We publicly release our code at \href{https://github.com/Mubuky/Self-Foveate}{https://github.com/Mubuky/Self-Foveate}
\end{abstract}

\begin{figure}[t!]
    \centering
    \includegraphics[width=1.0\columnwidth,trim=0.0cm 0.0cm 0.0cm 0.0cm]{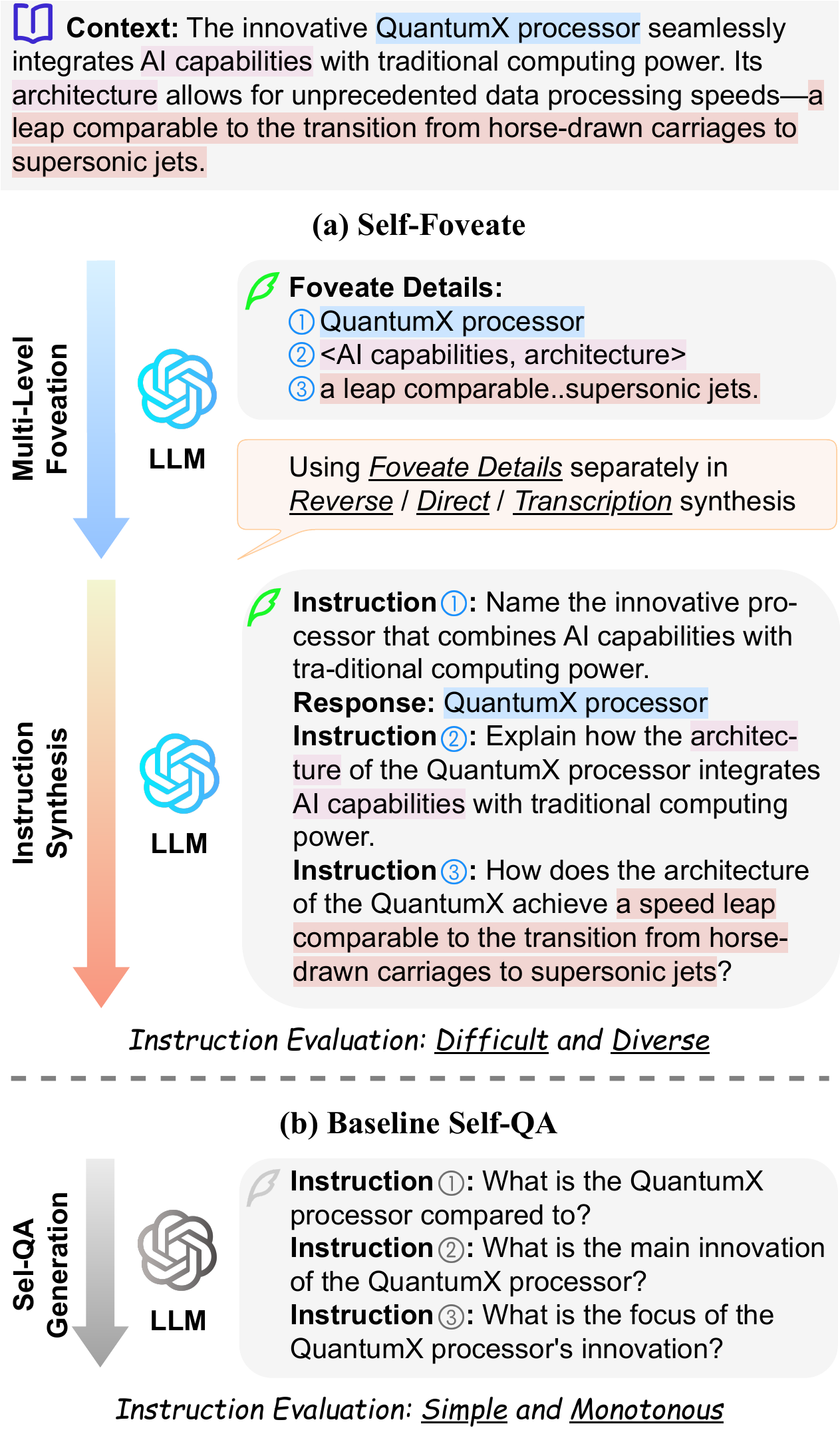}
    \caption{Illustration of (a) \approach in contrast with (b) Baseline Self-QA. For \approach, the multi-level foveation enables the LLM to extract details (highlighted in distinct colors) of the text, subsequently synthesizing instructions with diversity and difficulty via distinct synthesis paradigms. In comparison, Self-QA employs single-step generation that produces simple and monotonous instruction candidates.}
    \label{fig:small}
    \vspace{-1em}
\end{figure}

\begin{table*}[t!]
    \small
    \centering
    \renewcommand\arraystretch{1.2}
    \setlength{\tabcolsep}{4pt}
    \adjustbox{max width=\textwidth}{%
        \begin{tabular}{l c >{\centering}p{2.0cm} >{\centering}p{2.0cm} >{\centering}p{2.0cm} >{\centering\arraybackslash}p{2.0cm}}
            \toprule[1pt]
            \textbf{Related} & \textbf{Data} & \textbf{\textit{w/o} Human} & \textbf{Correctness} & \textbf{Diversity} & \textbf{Difficulty} \\
            \textbf{Work} & \textbf{Utilization} & \textbf{Annotation} & \textbf{Guarantee} & \textbf{Augment} & \textbf{Augment} \\
            \midrule
            \rowcolor{gray!10}
            Self-Instruct \citep{wang2023selfinstructaligninglanguagemodels} & Seed QA examples & \xmarkcolor & \xmarkcolor & \checkmarkcolor & \xmarkcolor \\
            Self-Align \citep{sun2023principledrivenselfalignmentlanguagemodels} & Seed QA examples & \xmarkcolor & \xmarkcolor & \checkmarkcolor & \checkmarkerror \\
            \rowcolor{gray!10}
            Self-Chat \citep{xu2023baizeopensourcechatmodel} & Dialogue & \checkmarkcolor & \xmarkcolor & \xmarkcolor & \xmarkcolor \\
            Self-QA \citep{zhang2023selfqaunsupervisedknowledgeguided} & Unsupervised Knowledge & \checkmarkcolor & \checkmarkcolor & \checkmarkerror & \xmarkcolor \\
            \rowcolor{gray!10}
            LongForm \citep{köksal2024longformeffectiveinstructiontuning} & Web dataset & \checkmarkcolor & \checkmarkcolor & \checkmarkcolor & \xmarkcolor \\
            Humpback \citep{li2024selfalignmentinstructionbacktranslation} & Web dataset & \xmarkcolor & \checkmarkcolor & \xmarkcolor & \checkmarkerror \\
            \rowcolor{gray!10}
            ISARA \citep{guo2024humaninstructionfreellmselfalignmentlimited} & Seed QA examples & \checkmarkcolor & \checkmarkcolor & \checkmarkcolor & \xmarkcolor \\
            Wiki2023 \citep{jiang-etal-2024-instruction} & Unsupervised Text & \checkmarkerror & \checkmarkcolor & \xmarkcolor  & \xmarkcolor \\\hdashline
            \rowcolor{gray!10}
            \textbf{\approach (Ours)} & Unsupervised Text & \checkmarkcolor & \checkmarkcolor & \checkmarkcolor & \checkmarkcolor \\
            \bottomrule[1pt]
        \end{tabular}
    }
    \caption{A comparative analysis of various task generation methodologies or frameworks. The gray checkmark symbol denotes that the work may partially accomplish specific objectives (though not comprehensively). The red cross marker indicates either the work's failure to achieve the stated objective or absence of explicit documentation regarding this goal.}
    \label{tab:comparison}
\end{table*}

\section{Introduction}
Large language models (LLMs), such as GPT-4o \citep{openai2024gpt4technicalreport}, Claude 3.5 Sonnet \citep{Claude}, and Llama 3.1 \citep{grattafiori2024llama3herdmodels}, have demonstrated exceptional instruction-following capabilities \citep{zhou2023instructionfollowingevaluationlargelanguage} and increasingly strong problem-solving abilities \citep{cobbe2021gsm8k, Dua2019DROP}. A critical component in training such models involves fine-tuning with extensive supervised question-answering instruction data \citep{wang2023aligninglargelanguagemodels}. However, substantial challenges persist in constructing large-scale, high-quality supervised fine-tuning (SFT) data for instruction tuning \citep{zhang2024instructiontuninglargelanguage}. The prohibitive costs of human annotation \citep{wang2023selfinstructaligninglanguagemodels, sun2023principledrivenselfalignmentlanguagemodels, xu2023baizeopensourcechatmodel}, coupled with difficulties in ensuring data diversity and quality control \citep{ge2024clusteringrankingdiversitypreservedinstruction}, continue to impede technological advancements.

Given the strong instruction-following and generative capabilities of modern LLMs, researchers are actively exploring methodologies to leverage these models for synthetic data synthesis \citep{wang2023selfinstructaligninglanguagemodels, zhang2023selfqaunsupervisedknowledgeguided, nayak2024learninggenerateinstructiontuning, wu2024unigenunifiedframeworktextual}. The primary objectives are to produce high-quality datasets with enhanced diversity and difficulty, thereby improving the performance of fine-tuned models on downstream tasks.

A promising recent paradigm in instruction data synthesis is \textit{unsupervised text-based instruction synthesis}. The advantages of this paradigm become particularly pronounced when handling massive unsupervised text corpora, a ubiquitous resource containing rich world knowledge and linguistic patterns. By leveraging LLMs' intrinsic capabilities in contextual understanding and logical reasoning, this paradigm eliminates the need for manual annotation while grounding instruction synthesis in the given unsupervised textual materials.

Recent automated methods such as Self-QA \citep{zhang2023selfqaunsupervisedknowledgeguided} have significantly optimized the data generation pipeline while reducing human labor costs. However, as shown in \autoref{tab:comparison}, critical challenges remain unresolved:
\textbf{(1) Diversity of Instructions:} While existing frameworks continuously refine data generation strategies and enhance post-synthesis filtering for instruction data, limitations persist in the diversity of synthesized instructions, which tend to exhibit repetitive patterns in both structure and topic coverage, potentially constraining the generalization of fine-tuned models.
\textbf{(2) Difficulty of Instructions:} Current synthesis methodologies generally lack emphasis on instruction complexity and depth. For instance, Self-QA \citep{zhang2023selfqaunsupervisedknowledgeguided} directly acquires instructions through single-step generation without guaranteed difficulty levels (as illustrated in \autoref{fig:small}). Synthesized instructions often exhibit simplistic structures and fail to capture complex inter-entity relationships, limiting models' ability to produce nuanced responses.
These limitations suggest that existing methods fail to fully exploit the rich, multi-granularity information inherent in unsupervised text, leaving substantial potential untapped.

To address these challenges, we observe that textual data inherently contains abundant detailed information encompassing entities (\textit{e.g.}, \textit{QuantumX processor}), attributes, relations, and writing techniques. For instance, the statement \textit{``In the realm of technology, speed is not just a luxury''} implicitly frames speed as a fundamental necessity through \textit{subtle analogy} (as multi-color annotated in \autoref{fig:small}). This rich, multi-granularity information remains underutilized in existing methods.

This multi-level nature of textual information finds a compelling parallel in human visual perception. In the human visual system, the fovea centralis, a small region of the retina with the highest density of photoreceptor cells, enables sharp central vision for resolving fine details at the point of fixation \citep{wandell1995foundations}. Beyond this focal point, rapid saccadic eye movements shift the fovea across different regions of interest, and the brain integrates information from these dispersed fixation points to comprehend spatial relationships \citep{findlay2003active}. Meanwhile, peripheral vision captures the global structure and holistic patterns of the scene at a coarser resolution. This multi-level visual processing, spanning fine-grained foveal detail, cross-region saccadic integration, and holistic peripheral awareness, inspires our approach to multi-granularity text comprehension.

Motivated by these textual observations and the analogous multi-level processing in human foveation, this paper proposes \approach, a comprehensive LLM-based methodology designed to automatically synthesize instructions from unsupervised text. As illustrated in \autoref{fig:small}, \approach introduces a ``Micro-Scatter-Macro'' multi-level foveation methodology in which \textit{micro-foveate} mirrors foveal fixation on fine-grained details, \textit{scatter-foveate} emulates saccadic integration across dispersed information, and \textit{macro-foveate} parallels peripheral perception of holistic patterns, thereby systematically extracting detailed information from raw text and subsequently synthesizing instructions with enhanced diversity and difficulty through three synthesis paradigms. Furthermore, \approach integrates a re-synthesis module to improve both the fidelity of instructions to source text and their overall quality.

To summarize, the key contributions of this paper are as follows:
\begin{itemize}[nolistsep, leftmargin=*]
\item[$\triangleright$] We systematically analyze unsupervised text-based instruction synthesis, identifying critical limitations of existing methods in both instruction diversity and difficulty.
\item[$\triangleright$] We propose \approach, a multi-level foveation methodology inspired by human visual perception that synthesizes diverse and challenging instructions from unsupervised text via three complementary strategies: micro-, scatter-, and macro-foveation.
\item[$\triangleright$] We conduct extensive experiments covering diversity and difficulty analysis, downstream task evaluation, and data scaling analysis. Results demonstrate that \approach consistently outperforms existing methods in unsupervised text-based instruction synthesis.
\end{itemize}

\begin{figure*}[t]
    \centering
    \includegraphics[width=1.0\textwidth,trim=1.0cm 3.7cm 3.7cm 1.0cm]{figures/big_pic.pdf}
    \caption{The \approach workflow is designed for instruction synthesis based on unsupervised text. \approach takes unsupervised text as input, extracts foveate elements, foveate groups, and foveate segments, then synthesizes instruction tuning data through these extracted details.}
    \label{fig:big}
    \vspace{-0.5em}
\end{figure*}

\section{Related Work}

\paragraph{Instruction Data Synthesis}
Multitask instruction fine-tuning
\citep{wei2022finetunedlanguagemodelszeroshot} substantially improves language
models' ability to follow instructions and generalize to unseen tasks
\citep{sanh2022multitaskpromptedtrainingenables, mishra-etal-2022-cross,
chung2022scalinginstructionfinetunedlanguagemodels,
longpre2023flancollectiondesigningdata, zhou2023limaalignment,
li2024selfalignmentinstructionbacktranslation}. As manually curating
large-scale instruction data remains costly
\citep{liu2024bestpracticeslessonslearned} and the constraints that data
availability imposes on model capability continue to grow
\citep{villalobos2024rundatalimitsllm,
long2024llmsdrivensyntheticdatageneration}, recent work leverages LLMs to
synthesize instruction--response pairs automatically. Self-Instruct
\citep{wang2023selfinstructaligninglanguagemodels} bootstraps new tasks from a
small set of seed demonstrations; Unnatural Instructions
\citep{honovich-etal-2023-unnatural} and Alpaca \citep{alpaca} scale this idea
with minimal human intervention; and Self-QA
\citep{zhang2023selfqaunsupervisedknowledgeguided} further removes the need for
seed tasks by grounding synthesis in unsupervised text. Other efforts refine
this pipeline through GPT-4 distillation \citep{peng2023instructiontuninggpt4},
self-rewarding mechanisms \citep{yuan2024selfrewardinglanguagemodels}, and
generalized synthesis frameworks \citep{li2024syntheticdataalmostscratch}. Our
work follows the unsupervised text-based synthesis paradigm, focusing
specifically on the quality dimensions of diversity and difficulty that prior
pipelines leave underaddressed.

\paragraph{Diversity and Difficulty of Synthesized Instructions}
Despite the rapid progress in instruction data synthesis, two persistent
challenges remain underexplored. First, synthesized instructions often lack
variety in both form and subject matter, narrowing the \textit{diversity} of
the resulting datasets \citep{chung-etal-2023-increasing,
ge2024clusteringrankingdiversitypreservedinstruction}. Second, most synthesis
pipelines lack explicit mechanisms to control instruction \textit{difficulty};
for example, Self-QA \citep{zhang2023selfqaunsupervisedknowledgeguided}
produces instructions in a single pass, offering no explicit control over their
complexity. While earlier approaches based on smaller pretrained language models
\citep{schick2021generatingdatasetspretrainedlanguage} were constrained by
limited model capacity, even modern LLM-based pipelines have not fully
addressed these quality dimensions. A common limitation is that existing methods
treat source text as a monolithic input rather than systematically exploiting
its multi-granularity structure, leaving substantial room for more varied and
challenging instructions. To close this gap, \approach introduces a multi-level
foveation methodology that guides the extraction of textual information at
three complementary granularities and yields instructions of greater variety
and complexity, all without human annotation.

\section{\approach}
\label{sec:method}
We propose \approach, denoted as \(\mathcal{F}\), a multi-level method for synthesizing instructions from unsupervised text without human-annotated samples. \approach consists of three levels and a re-synthesis module. Formally, given an unsupervised text set \(\mathcal{D}\), \approach transforms each document \(d_i \in \mathcal{D}\) through a set of synthesis paths \(\{\mathcal{F}_j\}\). The final synthesized dataset, \(\mathcal{D}_{\text{gen}}\), is the union of subsets produced by applying every synthesis path to every document:
\begin{equation}
    \label{eq:dataset_union}
    \mathcal{D}_{\text{gen}} = \mathcal{F}(\mathcal{D}) = \bigcup_{d_i \in \mathcal{D}} \bigcup_{\mathcal{F}_j\in \mathcal{F}} \mathcal{F}_j(d_i)
\end{equation}
The synthesis process is designed to maximize both the diversity and difficulty of the resulting instructions.

As shown in \autoref{fig:big}, \approach incorporates three levels and one module:
\begin{itemize}[nolistsep, leftmargin=*]
\item[$\triangleright$] \textbf{Micro-foveate Level:} Extracts fine-grained \textit{foveate elements} (entities and attributes) from unsupervised text, then uses each element as a seed answer to synthesize instructions via reverse synthesis (\S\ref{subsec:sf_micro}).
\item[$\triangleright$] \textbf{Scatter-foveate Level:} Broadly extracts foveate elements and combines them into \textit{foveate groups}, then applies direct synthesis to produce instructions requiring cross-entity reasoning (\S\ref{subsec:sf_scatter}).
\item[$\triangleright$] \textbf{Macro-foveate Level:} Identifies text segments employing rhetorical or figurative devices as \textit{foveate segments}, then converts them into instructional form via transcription synthesis (\S\ref{subsec:sf_macro}).
\item[$\triangleright$] \textbf{Re-synthesis Module:} Filters the synthesized instruction set for unanswerable instructions and iteratively re-synthesizes them using successfully synthesized examples as references, improving fidelity to the source text (\S\ref{subsec:sf_re}).
\end{itemize}

\subsection{Micro-foveate Level}
\label{subsec:sf_micro}
Unsupervised text typically contains a hierarchy of information: primary entities, secondary entities, and their respective attributes. Without explicit guidance, a teacher LLM synthesizing instructions from such text tends to focus on salient surface-level content, overlooking secondary entities and fine-grained attributes. The resulting instruction sets therefore miss critical information, limiting downstream model performance. The micro-foveate level addresses this gap through a fine-grained extraction mechanism paired with reverse synthesis, ensuring that individually important pieces of information are preserved in the synthesized instruction sets. Covering a broader range of entities and attributes enhances diversity, while probing specific entity attributes increases difficulty.
\paragraph{Micro-Foveate Mechanism} This mechanism aims to guide the teacher LLM to synthesize instructions from a fine-grained perspective of the unsupervised text. We introduce the concept of ``\textit{foveate elements}'', which broadly encompass all entities and their attributes within a given text. We guide the teacher LLM to extract more foveate elements than needed and then retain those whose embeddings have the highest cosine similarity to the full text embedding, filtering out extractions that are tangential to the source content.
\paragraph{Reverse Synthesis} Based on the selected foveate elements, we employ the reverse synthesis method to produce instructions. Specifically, we treat each foveate element as a potential answer to an instruction and guide the teacher LLM to synthesize a corresponding instruction from the unsupervised text, then derive the final answer. Even though the foveate elements can already serve as answers or parts of answers, to enhance the fluency, completeness, and accuracy of the answers, we choose to derive the answers anew in the reverse synthesis step rather than directly using the foveate elements as answers.

\subsection{Scatter-foveate Level}
\label{subsec:sf_scatter}
Beyond individual entities, unsupervised text often encodes implicit relationships (causal links, comparisons, or temporal dependencies) between entities or attributes that are scattered across different passages. A teacher LLM synthesizing instructions without targeted guidance is unlikely to surface these cross-entity connections, producing instruction sets that remain shallow. The scatter-foveate level tackles this by combining dispersed key information through a grouping mechanism and applying direct synthesis, so that the resulting instructions require reasoning about inter-entity relationships. This design broadens diversity by surfacing less obvious implicit information and raises difficulty by demanding cross-entity reasoning.
\paragraph{Scatter-Foveate Mechanism} This mechanism encourages the teacher LLM to reason about connections across different parts of the unsupervised text. We extract foveate elements more broadly from the text and randomly combine them into \textit{foveate groups}, with group sizes sampled from an empirical distribution.
\paragraph{Direct Synthesis} Based on the formed foveate groups, we synthesize instructions using the direct synthesis method, which treats each element in the foveate group as an indispensable component of the target instruction, guiding the teacher LLM to uncover implicit semantic connections between different entities or attributes and to embed such cross-entity reasoning in the synthesized instructions. After the instructions are synthesized, the corresponding answers are derived from the text.

\subsection{Macro-foveate Level}
\label{subsec:sf_macro}
At the broadest granularity, unsupervised text frequently employs figurative language and rhetorical devices (metaphor, hyperbole, contrastive foil, rhetorical questioning, and citation) that convey meaning beyond their literal content. A teacher LLM may overlook these devices without explicit prompting, missing the deeper communicative intent they carry. The macro-foveate level captures this holistic layer by identifying and extracting text segments that employ such writing techniques, then converting them into instructional form through transcription synthesis, ensuring that literary and rhetorical nuances are reflected in the synthesized instruction sets.
\paragraph{Macro-Foveate Mechanism} This mechanism directs the teacher LLM to comprehend unsupervised text from a holistic perspective. We highlight and extract the text segments employing such writing techniques, which are designated as \textit{foveate segments}.
\paragraph{Transcription Synthesis} Based on the identified foveate segments, we employ transcription synthesis to convert each one into an instructional format. This process transforms declarative foveate segments into interrogative or imperative forms. Subsequently, corresponding answers are synthesized according to the content of the unsupervised text.

\subsection{Re-synthesis Module}
\label{subsec:sf_re}
Due to limitations of the teacher LLM, not all instructions in a directly synthesized instruction set may be answerable from the source text. To improve fidelity, we employ a re-synthesis module that iteratively re-synthesizes these unanswerable instructions over multiple rounds, replacing them with instructions that can be fully answered based on the same text.
\paragraph{Single-Sample Reference Synthesis}
During re-synthesis, we process one failed instruction at a time and randomly sample a different subset of successfully synthesized instructions as reference examples in each iteration, improving the success rate by exposing the teacher LLM to varied reference contexts.
\paragraph{Hyperparameter Configuration}
The outputs of LLMs are significantly influenced by hyperparameters such as temperature, top-p, and frequency penalty. To improve re-synthesis success rates, we tune these hyperparameters to define a high-creativity configuration, enabling the teacher LLM to synthesize more varied instructions based on reference samples.

\begin{table}[b]
    \small
    \centering
    \renewcommand\arraystretch{1.2}
    \setlength{\tabcolsep}{4pt}
    \adjustbox{max width=\columnwidth}{
        \begin{tabular}{lccr}
            \toprule[1pt]
            \multirow{2}{*}{\textbf{Dataset}} & \multicolumn{2}{c}{\textbf{Source}} & \multirow{2}{*}{\textbf{\# Test Examples}} \\
            \cmidrule(lr){2-3}
            & \texttt{Question} & \texttt{Context} & \\
            \midrule
            SQuAD & Crowdsourced & Wikipedia & 11639 \\
            HotpotQA & Crowdsourced & Wikipedia & 2500 \\
            FilmWiki & LLM & Wikipedia & 7398 \\
            \bottomrule[1pt]
        \end{tabular}
    }
    \caption{Statistics for the evaluation datasets from our experiments.}
    \label{tab:dataset}
\end{table}

\begin{table}[t]
    \small
    \centering
    \renewcommand\arraystretch{1.2}
    \setlength{\tabcolsep}{4pt}
    \adjustbox{max width=\columnwidth}{
        \begin{tabular}{lccc}
            \toprule[1pt]
            \multirow{2}{*}{\textbf{Datasets}} & \multirow{2}{*}{\textbf{Methods}} & \multicolumn{2}{c}{\textbf{Diversity Metrics}} \\
            \cmidrule(lr){3-4}
            & & SelfBLEU Div. & Embedding Div. \\
            \midrule
            \multirow{5}{*}{SQuAD}
            & Self-QA & 0.593 & 0.838 \\
            & Bonito & 0.494 & 0.838 \\
            & Wiki2023 & 0.550 & 0.842 \\
            & \textbf{\approach} & \textbf{0.665} & \textbf{0.851} \\
            \cmidrule(lr){2-4}
            & Test Questions & 0.695 & 0.840 \\
            \midrule
            \multirow{5}{*}{HotpotQA}
            & Self-QA & 0.463 & 0.823 \\
            & Bonito & 0.371 & 0.769 \\
            & Wiki2023 & 0.554 & 0.822 \\
            & \textbf{\approach} & \textbf{0.607} & \textbf{0.835} \\
            \cmidrule(lr){2-4}
            & Test Questions & 0.634 & 0.786 \\
            \midrule
            \multirow{5}{*}{FilmWiki}
            & Self-QA & 0.406 & 0.687 \\
            & Bonito & 0.197 & 0.677 \\
            & Wiki2023 & 0.341 & 0.664 \\
            & \textbf{\approach} & \textbf{0.563} & \textbf{0.706} \\
            \cmidrule(lr){2-4}
            & Test Questions & 0.316 & 0.618 \\
            \bottomrule[1pt]
        \end{tabular}
    }
    \caption{Comparison of diversity metrics across different methods and datasets. The table presents SelfBLEU Diversity (SelfBLEU Div.) and Embedding Diversity (Embedding Div.) scores for various methods on the datasets. The diversity of the test questions from each dataset is also provided as a reference.}
    \label{tab:diversity}
\end{table}

\begin{table*}[t!]
    \small
    \centering
    \renewcommand\arraystretch{1.2}
    \setlength{\tabcolsep}{4pt}
    \adjustbox{max width=\textwidth}{
        \begin{tabular}{llcccccccccccc}
            \toprule[1pt]
            \multirow{3}{*}{\textbf{Model}} & \multirow{3}{*}{\textbf{Settings}} & \multicolumn{6}{c}{\textbf{GPT-4o mini}} & \multicolumn{6}{c}{\textbf{DeepSeek-V3}} \\
            \cmidrule(lr){3-8}
            \cmidrule(lr){9-14}
            & & \multicolumn{2}{c}{\textbf{SQuAD}} & \multicolumn{2}{c}{\textbf{HotpotQA}} & \multicolumn{2}{c}{\textbf{FilmWiki}} & \multicolumn{2}{c}{\textbf{SQuAD}} & \multicolumn{2}{c}{\textbf{HotpotQA}} & \multicolumn{2}{c}{\textbf{FilmWiki}} \\
            \cmidrule(lr){3-4}
            \cmidrule(lr){5-6}
            \cmidrule(lr){7-8}
            \cmidrule(lr){9-10}
            \cmidrule(lr){11-12}
            \cmidrule(lr){13-14}
            & & \texttt{Rec.} & \texttt{Acc.}
            & \texttt{Rec.} & \texttt{Acc.}
            & \texttt{Rec.} & \texttt{Acc.}
            & \texttt{Rec.} & \texttt{Acc.}
            & \texttt{Rec.} & \texttt{Acc.}
            & \texttt{Rec.} & \texttt{Acc.} \\
            \midrule
            \multirow{5}{*}{Llama-3.1-8B}
            & None* & 0.309 & 0.202 & 0.244 & 0.160 & 0.212 & 0.082 & 0.309 & 0.202 & 0.244 & 0.160 & 0.212 & 0.082 \\
            & Self-QA & 0.367 & 0.384 & 0.372 & 0.358 & 0.328 & 0.201 & 0.389 & 0.412 & 0.399 & 0.378 & 0.370 & 0.239 \\
            & Wiki2023 & 0.327 & 0.361 & 0.338 & 0.322 & 0.333 & 0.235 & 0.342 & 0.370 & 0.340 & 0.328 & 0.349 & 0.244 \\
            & Bonito* & 0.386 & 0.405 & 0.360 & 0.372 & 0.219 & 0.153 & 0.386 & 0.405 & 0.360 & 0.372 & 0.219 & 0.153 \\
            & \textbf{\approach} & \textbf{0.484} & \textbf{0.490} & \textbf{0.507} & \textbf{0.486} & \textbf{0.512} & \textbf{0.367} & \textbf{0.481} & \textbf{0.491} & \textbf{0.525} & \textbf{0.501} & \textbf{0.548} & \textbf{0.397} \\
            \midrule
            \multirow{5}{*}{Qwen2.5-7B}
            & None* & 0.251 & 0.300 & 0.266 & 0.234 & 0.139 & 0.032 & 0.251 & 0.300 & 0.266 & 0.234 & 0.139 & 0.032 \\
            & Self-QA & 0.249 & 0.232 & 0.276 & 0.246 & 0.206 & 0.082 & 0.119 & 0.125 & 0.102 & 0.106 & 0.111 & 0.056 \\
            & Wiki2023 & 0.215 & 0.221 & 0.135 & 0.112 & 0.192 & 0.093 & 0.170 & 0.083 & 0.197 & 0.203 & 0.202 & 0.136 \\
            & Bonito* & 0.143 & 0.109 & 0.212 & 0.199 & 0.168 & 0.098 & 0.143 & 0.109 & 0.212 & 0.199 & 0.168 & 0.098 \\
            & \textbf{\approach} & \textbf{0.408} & \textbf{0.414} & \textbf{0.372} & \textbf{0.329} & \textbf{0.283} & \textbf{0.140} & \textbf{0.388} & \textbf{0.389} & \textbf{0.342} & \textbf{0.331} & \textbf{0.261} & \textbf{0.140} \\
            \midrule
            \multirow{5}{*}{Gemma-2-9B}
            & None* & 0.224 & 0.121 & 0.175 & 0.078 & 0.211 & 0.099 & 0.224 & 0.121 & 0.175 & 0.078 & 0.221 & 0.099 \\
            & Self-QA & 0.383 & 0.409 & 0.408 & 0.389 & 0.429 & 0.315 & 0.402 & 0.435 & 0.424 & 0.408 & 0.509 & 0.386 \\
            & Wiki2023 & 0.336 & 0.378 & 0.361 & 0.352 & 0.478 & 0.384 & 0.364 & 0.399 & 0.373 & 0.365 & 0.494 & 0.401 \\
            & Bonito* & 0.411 & 0.457 & 0.366 & 0.373 & 0.255 & 0.196 & 0.411 & 0.457 & 0.366 & 0.373 & 0.255 & 0.196 \\
            & \textbf{\approach} & \textbf{0.507} & \textbf{0.525} & \textbf{0.537} & \textbf{0.520} & \textbf{0.672} & \textbf{0.528} & \textbf{0.499} & \textbf{0.514} & \textbf{0.552} & \textbf{0.525} & \textbf{0.697} & \textbf{0.581} \\
            \bottomrule[1pt]
        \end{tabular}
    }
    \caption{Recall (Rec.) and LLM Accuracy (Acc.) on downstream tasks: \approach vs. baselines. Results include models fine-tuned with instructions synthesized by GPT-4o mini or DeepSeek-V3, as well as reference non-instruction-tuned models (None). * Indicates that the base model was not fine-tuned using instructions synthesized by GPT-4o mini or DeepSeek-V3.}
    \label{tab:downstream}
\end{table*}

\section{Experiment}
\label{sec:experiment}
Our experiments focus on three critical research questions:
(Q1) How effective is \approach in enhancing instruction diversity?
(Q2) How effective is \approach in enhancing instruction difficulty?
(Q3) How effective is \approach in improving the model's problem-solving capabilities during instruction fine-tuning?

\subsection{Experimental Setup}
\label{subsec:setup}
\paragraph{Datasets}\label{par:datasets} We employ three independent datasets, including the training set of the FilmWiki dataset containing 2,385 unsupervised texts with corresponding question-answer pairs. Additionally, we sample two widely used benchmark QA datasets from the MRQA 2019 shared task~\citep{fisch-etal-2019-mrqa}: SQuAD~\citep{rajpurkar-etal-2016-squad} (following Bonito~\citep{nayak-etal-2024-learning}) and HotpotQA~\citep{yang-etal-2018-hotpotqa}. To maintain comparable data scale and computational costs to the FilmWiki dataset, we extract 2,500 unsupervised texts with corresponding QA pairs from each dataset's training split as substitutes for the complete collections. Further implementation details are provided in \autoref{tab:dataset}.
\paragraph{Baselines}\label{par:baselines} We consider four baselines: zero-shot, Self-QA, Bonito, and Wiki2023. For the zero-shot baseline, we prompt models for evaluation without leveraging any unsupervised texts from the target task (\textbf{None}). The Self-QA baseline employs an unsupervised knowledge-guided method for extracting instruction-question-answer triples (\textbf{Self-QA})~\citep{zhang2023selfqaunsupervisedknowledgeguided}. The Bonito baseline utilizes a 7B-sized specialized model to generate various types of questions from unsupervised text. We configure it to produce questions of the ``question answering without choices'' type and obtain a sufficient number of samples through multiple sampling iterations (\textbf{Bonito}). The Wiki2023 baseline implements a text-based QA pair extraction methodology (\textbf{Wiki2023}). All baselines use the same unsupervised texts as our method, as specified in Section \ref{par:datasets}.
\paragraph{Instruction Synthesis}
As described in \autoref{sec:method}, we process unsupervised texts through \approach using GPT-4o mini and DeepSeek-V3 to synthesize instructional data. We implement two distinct hyperparameter configurations emphasizing high stability and high creativity, respectively, with detailed specifications in \autoref{sec:app_taskgen}.
\paragraph{Base Models}
We select three state-of-the-art open-source foundation models as our base models prior to instruction tuning: Meta-Llama-3.1-8B~\citep{grattafiori2024llama3herdmodels}, Qwen2.5-7B~\citep{qwen2.5}, and Gemma-2-9B~\citep{gemmateam2024gemma2improvingopen}. These decoder-only language models employ next-word prediction objectives and were pretrained on trillions of tokens without any instruction-based fine-tuning.

\subsection{Diversity Analysis}
\label{exp:diversity}
To explicitly evaluate the diversity of the instructions in a metric-driven manner, we follow established practices recommended by \citet{zhu2018texygenbenchmarkingplatformtext, perez2022redteaminglanguagemodels, tevet2021evaluatingevaluationdiversitynatural} and employ two metrics: SelfBLEU scores and Sentence-BERT embedding distances~\citep{reimers-gurevych-2019-sentence}. These metrics capture different facets of diversity. SelfBLEU measures surface-level textual diversity, while embedding distances measure semantic diversity. For SelfBLEU, we compute average scores using $n$-grams for $n \in \{2, 3, 4, 5\}$, following the approach suggested by \citet{zhu2018texygenbenchmarkingplatformtext}. We use the implementation of the SelfBLEU metric by \citet{alihosseini-etal-2019-jointly}. Further details are available in \autoref{sec:app_diversity}.

\autoref{tab:diversity} demonstrates that \approach-synthesized instructions achieve substantial diversity improvements in both textual and semantic dimensions, attaining or even surpassing the diversity level of crowdsourced test questions through a low-cost automated process.

\subsection{Difficulty Analysis}
\label{exp:difficulty}
To investigate the difficulty of the synthesized instructions, we design prompts and employ \sotallm to conduct a rigorous head-to-head comparison between the instructions synthesized by \approach and those synthesized by baseline methods. Specifically, for each unsupervised text in each dataset, we use two contrasting methods to synthesize an equal number of instructions. We then provide \sotallm with a set of instructions synthesized by both methods under the same unsupervised text, ensuring that the instruction sets are anonymous and their relative positions are randomized. Subsequently, we record \sotallm's judgments and calculate the win rate for \approach. The results of the head-to-head comparison, as shown in \autoref{tab:difficulty}, demonstrate the consistently higher difficulty of the instructions synthesized by \approach. Further details can be found in \autoref{sec:app_difficulty}.

\begin{figure*}[t!]
    \centering
    \begin{subfigure}[b]{0.49\textwidth}
        \centering
        \includegraphics[width=\linewidth,trim=0.3cm 1.5cm 2.0cm 2.0cm]{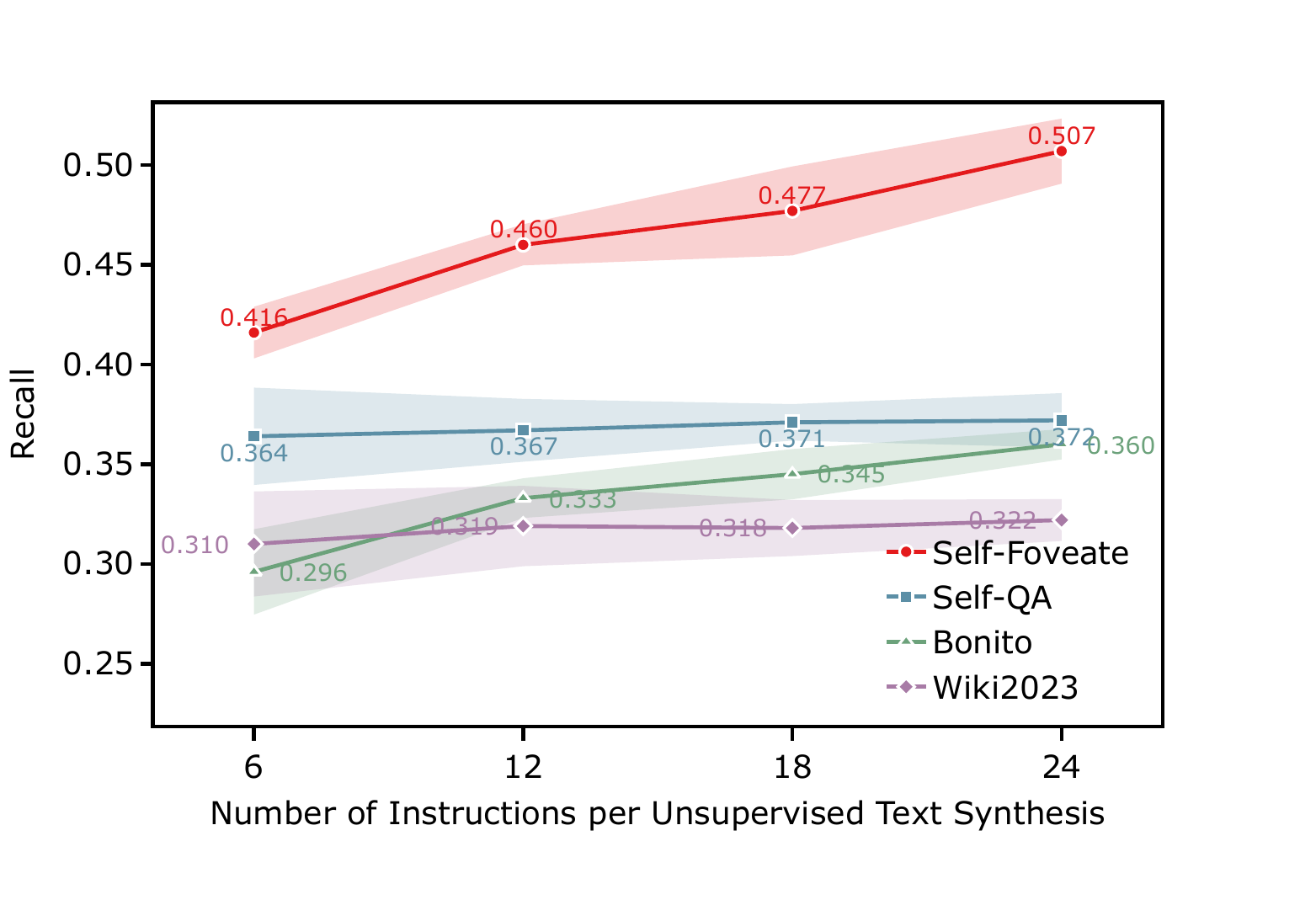}
    \end{subfigure}
    \hfill
    \begin{subfigure}[b]{0.49\textwidth}
        \centering
        \includegraphics[width=\linewidth,trim=0.3cm 1.5cm 2.0cm 2.0cm]{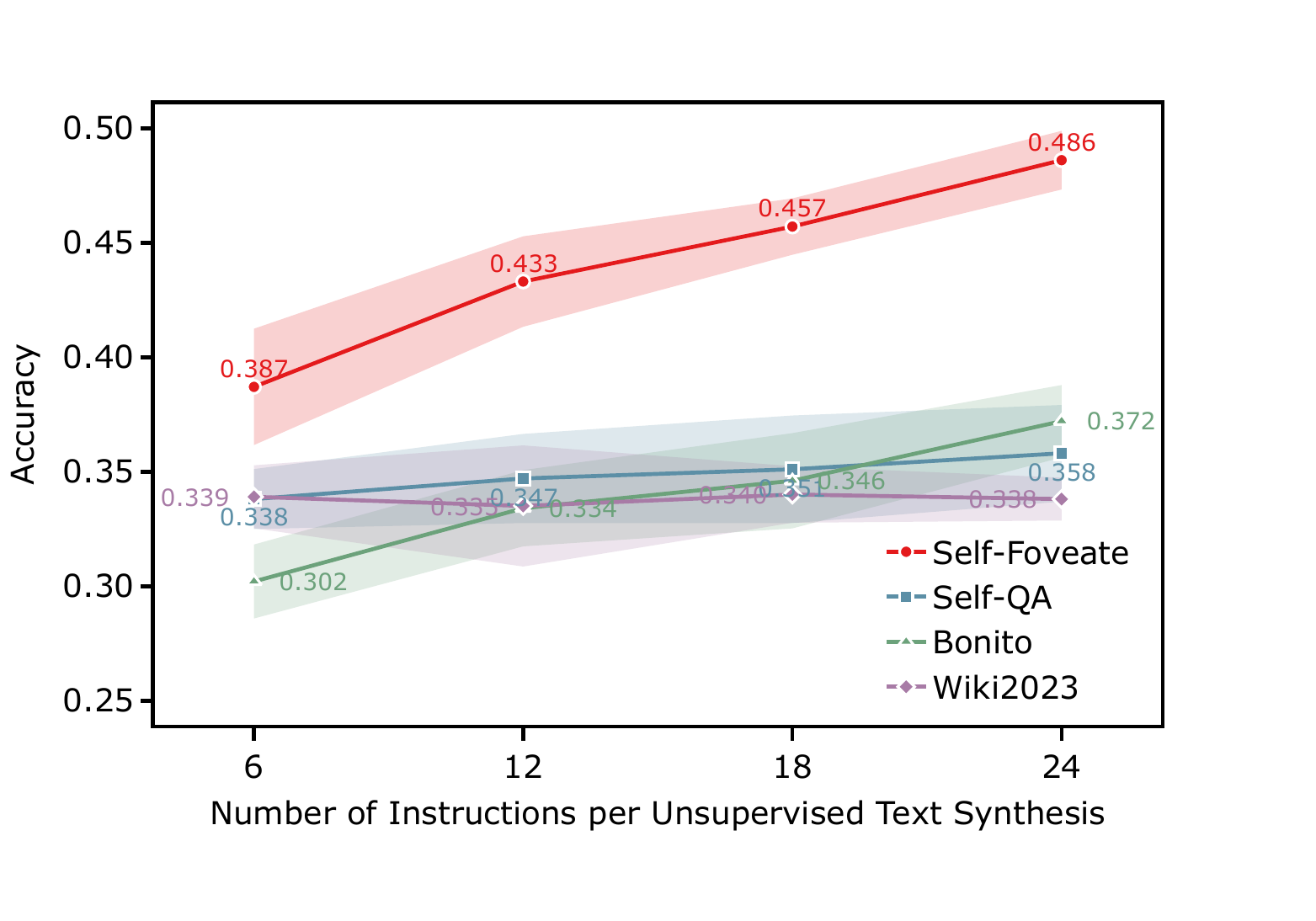}
    \end{subfigure}
    \caption{Impact of instruction set scale on model fine-tuning performance for \approach and baselines: (a) Recall and (b) LLM Accuracy.}
    \label{fig:trend}
    \vspace{-1em}
\end{figure*}

\begin{table}[b!]
    \small
    \centering
    \renewcommand\arraystretch{1.2}
    \setlength{\tabcolsep}{4pt}
    \adjustbox{max width=\columnwidth}{
        \begin{tabular}{p{1.8cm}p{1.9cm}cc}
            \toprule[1pt]
            \multirow{2}{*}{\textbf{Dataset}} & \multirow{2}{*}{\textbf{Baseline}} & \multicolumn{2}{c}{\textbf{\ \approach WR.\ }} \\
            \cmidrule(lr){3-4}
            & & \texttt{Win} & \texttt{Lose} \\
            \midrule
            \multirow{3}{*}{SQuAD} & Self-QA & \textbf{70.64\%} & 29.36\% \\
            & Wiki2023 & \textbf{80.83\%} & 19.17\% \\
            & Bonito & \textbf{99.96\%} & 00.04\% \\
            \midrule
            \multirow{3}{*}{HotpotQA} & Self-QA & \textbf{89.52\%} & 10.48\% \\
            & Wiki2023 & \textbf{91.17\%} & 08.83\% \\
            & Bonito & \textbf{100.00\%} & 00.00\% \\
            \midrule
            \multirow{3}{*}{FilmWiki} & Self-QA & \textbf{85.12\%} & 14.88\% \\
            & Wiki2023 & \textbf{95.08\%} & 04.92\% \\
            & Bonito & \textbf{96.31\%} & 03.69\% \\
            \bottomrule[1pt]
        \end{tabular}
    }
    \caption{Head-to-head comparison of instruction difficulty for \approach against baselines through win rates (WR.) across datasets.}
    \label{tab:difficulty}
\end{table}

\subsection{Problem-solving Capabilities}
\label{exp:problem}
To evaluate the impact of instruction-tuning datasets on model performance in downstream tasks, we fine-tune open-source models and assess their performance on in-distribution problems. \autoref{tab:downstream} illustrates the fine-tuning performance of open-source models using instruction sets synthesized by two teacher LLMs (GPT-4o mini and DeepSeek-V3), with equal quantities of instructions per unsupervised text. As answers in test problems often consist of phrase-level or sentence-fragment extractions from source texts while large language models (LLMs) tend to generate more comprehensive responses, we select both recall and LLM-evaluated accuracy as evaluation metrics. Higher recall indicates that generated answers are more likely to contain correct responses, while higher LLM-evaluated accuracy reflects stronger agreement between model-produced answers and ground truth labels, as judged by an LLM evaluator.

To control potential confounding effects from the number of instructions synthesized per unsupervised text and examine the agreement between these two metrics, we investigate the influence of the scale of instruction sets synthesized by \approach and baseline methods on model fine-tuning performance using the HotpotQA dataset, as shown in \autoref{fig:trend}. Additional details regarding downstream task evaluations are provided in \autoref{sec:app_downstream}.

\autoref{tab:downstream} demonstrates that all three base models achieve the best performance across downstream tasks in both metrics when fine-tuned with instructions synthesized via \approach, regardless of the teacher LLM used for synthesis. Notably, for Qwen2.5-7B, instruction data synthesized by baseline methods can lead to performance degradation compared to the base model without fine-tuning, further demonstrating the superiority of \approach in instruction synthesis. Our analysis in \autoref{fig:trend} reveals that as the number of instructions synthesized by \approach per unsupervised text increases, downstream performance improves significantly, with an expanding gap compared to baseline methods. \autoref{fig:trend} also confirms the consistency between recall and LLM-evaluated accuracy across varying instruction set scales.

\subsection{Ablation Studies}
\label{exp:ablation}
To validate the effectiveness and necessity of each component in \approach, we conduct comprehensive ablation studies focusing on two key aspects: (1) the contribution of each core component in the multi-level foveation framework, and (2) the impact of the answer regeneration mechanism in reverse synthesis.

\subsubsection{Component-wise Analysis}
We evaluate the necessity of each core component by systematically removing individual components from the complete \approach framework. \autoref{tab:ablation_components} presents the ablation results on the SQuAD dataset using Llama-3.1-8B as the base model.

\begin{table}[t]
    \small
    \centering
    \renewcommand\arraystretch{1.2}
    \setlength{\tabcolsep}{4pt}
    \adjustbox{max width=\columnwidth}{
        \begin{tabular}{lcc}
            \toprule[1pt]
            \textbf{Setting} & \textbf{Recall} & \textbf{LLM Acc.} \\
            \midrule
            w/o Micro-Foveate & 0.283 & 0.277 \\
            w/o Scatter-Foveate & 0.274 & 0.260 \\
            w/o Macro-Foveate & 0.344 & 0.339 \\
            \midrule
            \textbf{\approach (Full)} & \textbf{0.484} & \textbf{0.490} \\
            \bottomrule[1pt]
        \end{tabular}
    }
    \caption{Ablation study results showing the contribution of each core component in \approach on the SQuAD dataset with Llama-3.1-8B.}
    \label{tab:ablation_components}
\end{table}

The removal of any single component leads to clear performance degradation, with scatter-foveate removal causing the largest drop (Recall from 0.484 to 0.274), confirming that all three components are necessary and complementary.

\subsubsection{Answer Regeneration Analysis}
In our reverse synthesis paradigm, although foveate elements (e.g., ``QuantumX processor'') can directly serve as answers, we regenerate answers to enhance fluency, completeness, and semantic coherence. To validate this design choice, we compare the quality of answers with and without the regeneration mechanism. \autoref{tab:ablation_regeneration} presents the comparative analysis on the SQuAD dataset, evaluating both fluency and completeness of synthesized answers using GPT-4o.

\begin{table}[h]
    \small
    \centering
    \renewcommand\arraystretch{1.2}
    \setlength{\tabcolsep}{4pt}
    \adjustbox{max width=\columnwidth}{
        \begin{tabular}{lccc}
            \toprule[1pt]
            \textbf{Setting} & \textbf{High (\%)} & \textbf{Medium (\%)} & \textbf{Low (\%)} \\
            \midrule
            \rowcolor{gray!10}
            \multicolumn{4}{c}{\textit{Fluency}} \\
            w/o Regeneration & 64.9 & 19.5 & 15.6 \\
            Reverse Synthesis & \textbf{93.0} & \textbf{6.8} & \textbf{0.2} \\
            \midrule
            \rowcolor{gray!10}
            \multicolumn{4}{c}{\textit{Completeness}} \\
            w/o Regeneration & 36.7 & 27.5 & 35.7 \\
            Reverse Synthesis & \textbf{76.6} & \textbf{19.7} & \textbf{3.7} \\
            \bottomrule[1pt]
        \end{tabular}
    }
    \caption{Comparison of answer quality with and without regeneration mechanism in reverse synthesis on the SQuAD dataset. Fluency and completeness are evaluated on a three-point scale (High/Medium/Low).}
    \label{tab:ablation_regeneration}
\end{table}

The results show that answer regeneration substantially improves both fluency and completeness. This validates our choice to maintain consistency across all synthesis paradigms by synthesizing contextually appropriate and semantically coherent answers from the source text, rather than directly using extracted foveate elements.

\section{Conclusion}
\label{sec:conclusion}
In this paper, we presented \approach, an LLM-driven methodology for synthesizing diverse and challenging instructions from unsupervised text. Drawing inspiration from hierarchical human visual perception, \approach introduces a ``Micro-Scatter-Macro'' multi-level foveation methodology: \textit{micro-foveate} captures fine-grained entity details, \textit{scatter-foveate} integrates dispersed cross-entity relationships, and \textit{macro-foveate} extracts holistic rhetorical patterns. A re-synthesis module further enhances fidelity to source text and overall instruction quality. Comprehensive experiments across multiple unsupervised corpora and diverse model architectures demonstrate that \approach consistently outperforms existing methods in instruction diversity, difficulty, and downstream task performance. Our work establishes multi-level foveation as an effective paradigm for high-quality instruction synthesis from unsupervised text.

\section*{Acknowledgments}
\label{posec:acknowledgments}
This work was supported by the National Natural Science Foundation of China (NSFC) via grant 62441614 and 62176078. We thank the anonymous reviewers for their insightful comments and suggestions.

\section*{Limitations}
\label{posec:limitations}
Although \approach provides significant improvements in synthesizing diverse and difficulty-oriented fine-tuning instruction data based on unsupervised text, several limitations must be acknowledged. First, the computational cost of processing large-scale unsupervised text using closed-source state-of-the-art large language models (LLMs) remains substantial. Ensuring data quality requires multi-step reasoning and iterative processing on unsupervised text, which further escalates computational demands. Second, although \approach aims to fully synthesize data from unsupervised text and incorporates verification mechanisms, the inherent tendency of LLMs to hallucinate or generate erroneous information persists as a challenge. While precise prompting strategies and higher-quality unsupervised text can mitigate these inaccuracies, they cannot be entirely eliminated.

\section*{Ethical Consideration}
\label{posec:ethical}
We state that any research or application arising from this study is strictly authorized solely for research purposes. In our work, any unsupervised text datasets used are from public sources and do not contain any private information. In this paper, we have fully presented the prompts used by \approach in the Appendix. All synthesized instructions rely on the provided unsupervised text and are inspected by relevant modules. Therefore, our method strives to minimize potential safety and ethical risks as much as possible. However, during the process of synthesizing fine-tuning instruction data, maliciously provided unsupervised text can lead the model to produce harmful or inappropriate outputs, which is a shared problem. Additionally, potential unfairness and discrimination present in the unsupervised text might be amplified by LLMs during the instruction synthesis process. Ensuring the quality of synthesized fine-tuning instruction data in a safe and highly controllable manner is crucial. The application of these techniques should be guided by ethical considerations, with safeguards in place to prevent misuse and reduce the likelihood of producing harmful outcomes.

\bibliography{custom}

\begin{thebibliography}{55}
\providecommand{\natexlab}[1]{#1}

\bibitem[{Alihosseini et~al.(2019)Alihosseini, Montahaei, and
  Soleymani~Baghshah}]{alihosseini-etal-2019-jointly}
Danial Alihosseini, Ehsan Montahaei, and Mahdieh Soleymani~Baghshah. 2019.
\newblock \href {https://doi.org/10.18653/v1/W19-2311} {Jointly measuring
  diversity and quality in text generation models}.
\newblock In \emph{Proceedings of the Workshop on Methods for Optimizing and
  Evaluating Neural Language Generation}, pages 90--98, Minneapolis, Minnesota.
  Association for Computational Linguistics.

\bibitem[{Anthropic(2024)}]{Claude}
Anthropic. 2024.
\newblock Claude.
\newblock \url{https://www.anthropic.com/claude}.

\bibitem[{Chung et~al.(2022)Chung, Hou, Longpre, Zoph, Tay, Fedus, Li, Wang,
  Dehghani, Brahma, Webson, Gu, Dai, Suzgun, Chen, Chowdhery, Castro-Ros,
  Pellat, Robinson, Valter, Narang, Mishra, Yu, Zhao, Huang, Dai, Yu, Petrov,
  Chi, Dean, Devlin, Roberts, Zhou, Le, and
  Wei}]{chung2022scalinginstructionfinetunedlanguagemodels}
Hyung~Won Chung, Le~Hou, Shayne Longpre, Barret Zoph, Yi~Tay, William Fedus,
  Yunxuan Li, Xuezhi Wang, Mostafa Dehghani, Siddhartha Brahma, Albert Webson,
  Shixiang~Shane Gu, Zhuyun Dai, Mirac Suzgun, Xinyun Chen, Aakanksha
  Chowdhery, Alex Castro-Ros, Marie Pellat, Kevin Robinson, Dasha Valter,
  Sharan Narang, Gaurav Mishra, Adams Yu, Vincent Zhao, Yanping Huang, Andrew
  Dai, Hongkun Yu, Slav Petrov, Ed~H. Chi, Jeff Dean, Jacob Devlin, Adam
  Roberts, Denny Zhou, Quoc~V. Le, and Jason Wei. 2022.
\newblock \href {https://arxiv.org/abs/2210.11416} {Scaling
  instruction-finetuned language models}.
\newblock \emph{Preprint}, arXiv:2210.11416.

\bibitem[{Chung et~al.(2023)Chung, Kamar, and
  Amershi}]{chung-etal-2023-increasing}
John Chung, Ece Kamar, and Saleema Amershi. 2023.
\newblock \href {https://doi.org/10.18653/v1/2023.acl-long.34} {Increasing
  diversity while maintaining accuracy: Text data generation with large
  language models and human interventions}.
\newblock In \emph{Proceedings of the 61st Annual Meeting of the Association
  for Computational Linguistics (Volume 1: Long Papers)}, pages 575--593,
  Toronto, Canada. Association for Computational Linguistics.

\bibitem[{Cobbe et~al.(2021)Cobbe, Kosaraju, Bavarian, Chen, Jun, Kaiser,
  Plappert, Tworek, Hilton, Nakano, Hesse, and Schulman}]{cobbe2021gsm8k}
Karl Cobbe, Vineet Kosaraju, Mohammad Bavarian, Mark Chen, Heewoo Jun, Lukasz
  Kaiser, Matthias Plappert, Jerry Tworek, Jacob Hilton, Reiichiro Nakano,
  Christopher Hesse, and John Schulman. 2021.
\newblock Training verifiers to solve math word problems.
\newblock \emph{arXiv preprint arXiv:2110.14168}.

\bibitem[{Daniel~Han and team(2023)}]{unsloth}
Michael~Han Daniel~Han and Unsloth team. 2023.
\newblock \href {http://github.com/unslothai/unsloth} {Unsloth}.

\bibitem[{Dao(2024)}]{dao2023flashattention2}
Tri Dao. 2024.
\newblock Flash{A}ttention-2: Faster attention with better parallelism and work
  partitioning.
\newblock In \emph{International Conference on Learning Representations
  (ICLR)}.

\bibitem[{Dao et~al.(2022)Dao, Fu, Ermon, Rudra, and
  R{\'e}}]{dao2022flashattention}
Tri Dao, Daniel~Y. Fu, Stefano Ermon, Atri Rudra, and Christopher R{\'e}. 2022.
\newblock Flash{A}ttention: Fast and memory-efficient exact attention with
  {IO}-awareness.
\newblock In \emph{Advances in Neural Information Processing Systems
  (NeurIPS)}.

\bibitem[{DeepSeek-AI et~al.(2025)DeepSeek-AI, Liu, Feng, Xue, Wang, Wu, Lu,
  Zhao, Deng, Zhang, Ruan, Dai, Guo, Yang, Chen, Ji, Li, Lin, Dai, Luo, Hao,
  Chen, Li, Zhang, Bao, Xu, Wang, Zhang, Ding, Xin, Gao, Li, Qu, Cai, Liang,
  Guo, Ni, Li, Wang, Chen, Chen, Yuan, Qiu, Li, Song, Dong, Hu, Gao, Guan,
  Huang, Yu, Wang, Zhang, Xu, Xia, Zhao, Wang, Zhang, Li, Wang, Zhang, Zhang,
  Tang, Li, Tian, Huang, Wang, Zhang, Wang, Zhu, Chen, Du, Chen, Jin, Ge,
  Zhang, Pan, Wang, Xu, Zhang, Chen, Li, Lu, Zhou, Chen, Wu, Ye, Ye, Ma, Wang,
  Zhou, Yu, Zhou, Pan, Wang, Yun, Pei, Sun, Xiao, Zeng, Zhao, An, Liu, Liang,
  Gao, Yu, Zhang, Li, Jin, Wang, Bi, Liu, Wang, Shen, Chen, Zhang, Chen, Nie,
  Sun, Wang, Cheng, Liu, Xie, Liu, Yu, Song, Shan, Zhou, Yang, Li, Su, Lin, Li,
  Wang, Wei, Zhu, Zhang, Xu, Xu, Huang, Li, Zhao, Sun, Li, Wang, Yu, Zheng,
  Zhang, Shi, Xiong, He, Tang, Piao, Wang, Tan, Ma, Liu, Guo, Wu, Ou, Zhu,
  Wang, Gong, Zou, He, Zha, Xiong, Ma, Yan, Luo, You, Liu, Zhou, Wu, Ren, Ren,
  Sha, Fu, Xu, Huang, Zhang, Xie, Zhang, Hao, Gou, Ma, Yan, Shao, Xu, Wu,
  Zhang, Li, Gu, Zhu, Liu, Li, Xie, Song, Gao, and
  Pan}]{deepseekai2025deepseekv3technicalreport}
DeepSeek-AI, Aixin Liu, Bei Feng, Bing Xue, Bingxuan Wang, Bochao Wu, Chengda
  Lu, Chenggang Zhao, Chengqi Deng, Chenyu Zhang, Chong Ruan, Damai Dai, Daya
  Guo, Dejian Yang, Deli Chen, Dongjie Ji, Erhang Li, Fangyun Lin, Fucong Dai,
  Fuli Luo, Guangbo Hao, Guanting Chen, Guowei Li, H.~Zhang, Han Bao, Hanwei
  Xu, Haocheng Wang, Haowei Zhang, Honghui Ding, Huajian Xin, Huazuo Gao, Hui
  Li, Hui Qu, J.~L. Cai, Jian Liang, Jianzhong Guo, Jiaqi Ni, Jiashi Li, Jiawei
  Wang, Jin Chen, Jingchang Chen, Jingyang Yuan, Junjie Qiu, Junlong Li,
  Junxiao Song, Kai Dong, Kai Hu, Kaige Gao, Kang Guan, Kexin Huang, Kuai Yu,
  Lean Wang, Lecong Zhang, Lei Xu, Leyi Xia, Liang Zhao, Litong Wang, Liyue
  Zhang, Meng Li, Miaojun Wang, Mingchuan Zhang, Minghua Zhang, Minghui Tang,
  Mingming Li, Ning Tian, Panpan Huang, Peiyi Wang, Peng Zhang, Qiancheng Wang,
  Qihao Zhu, Qinyu Chen, Qiushi Du, R.~J. Chen, R.~L. Jin, Ruiqi Ge, Ruisong
  Zhang, Ruizhe Pan, Runji Wang, Runxin Xu, Ruoyu Zhang, Ruyi Chen, S.~S. Li,
  Shanghao Lu, Shangyan Zhou, Shanhuang Chen, Shaoqing Wu, Shengfeng Ye,
  Shengfeng Ye, Shirong Ma, Shiyu Wang, Shuang Zhou, Shuiping Yu, Shunfeng
  Zhou, Shuting Pan, T.~Wang, Tao Yun, Tian Pei, Tianyu Sun, W.~L. Xiao,
  Wangding Zeng, Wanjia Zhao, Wei An, Wen Liu, Wenfeng Liang, Wenjun Gao,
  Wenqin Yu, Wentao Zhang, X.~Q. Li, Xiangyue Jin, Xianzu Wang, Xiao Bi,
  Xiaodong Liu, Xiaohan Wang, Xiaojin Shen, Xiaokang Chen, Xiaokang Zhang,
  Xiaosha Chen, Xiaotao Nie, Xiaowen Sun, Xiaoxiang Wang, Xin Cheng, Xin Liu,
  Xin Xie, Xingchao Liu, Xingkai Yu, Xinnan Song, Xinxia Shan, Xinyi Zhou,
  Xinyu Yang, Xinyuan Li, Xuecheng Su, Xuheng Lin, Y.~K. Li, Y.~Q. Wang, Y.~X.
  Wei, Y.~X. Zhu, Yang Zhang, Yanhong Xu, Yanhong Xu, Yanping Huang, Yao Li,
  Yao Zhao, Yaofeng Sun, Yaohui Li, Yaohui Wang, Yi~Yu, Yi~Zheng, Yichao Zhang,
  Yifan Shi, Yiliang Xiong, Ying He, Ying Tang, Yishi Piao, Yisong Wang, Yixuan
  Tan, Yiyang Ma, Yiyuan Liu, Yongqiang Guo, Yu~Wu, Yuan Ou, Yuchen Zhu, Yuduan
  Wang, Yue Gong, Yuheng Zou, Yujia He, Yukun Zha, Yunfan Xiong, Yunxian Ma,
  Yuting Yan, Yuxiang Luo, Yuxiang You, Yuxuan Liu, Yuyang Zhou, Z.~F. Wu,
  Z.~Z. Ren, Zehui Ren, Zhangli Sha, Zhe Fu, Zhean Xu, Zhen Huang, Zhen Zhang,
  Zhenda Xie, Zhengyan Zhang, Zhewen Hao, Zhibin Gou, Zhicheng Ma, Zhigang Yan,
  Zhihong Shao, Zhipeng Xu, Zhiyu Wu, Zhongyu Zhang, Zhuoshu Li, Zihui Gu,
  Zijia Zhu, Zijun Liu, Zilin Li, Ziwei Xie, Ziyang Song, Ziyi Gao, and Zizheng
  Pan. 2025.
\newblock \href {https://arxiv.org/abs/2412.19437} {Deepseek-v3 technical
  report}.
\newblock \emph{Preprint}, arXiv:2412.19437.

\bibitem[{Dua et~al.(2019)Dua, Wang, Dasigi, Stanovsky, Singh, and
  Gardner}]{Dua2019DROP}
Dheeru Dua, Yizhong Wang, Pradeep Dasigi, Gabriel Stanovsky, Sameer Singh, and
  Matt Gardner. 2019.
\newblock {DROP}: A reading comprehension benchmark requiring discrete
  reasoning over paragraphs.
\newblock In \emph{Proc. of NAACL}.

\bibitem[{Findlay and Gilchrist(2003)}]{findlay2003active}
John~M. Findlay and Iain~D. Gilchrist. 2003.
\newblock \emph{Active Vision: The Psychology of Looking and Seeing}.
\newblock Oxford University Press, Oxford.

\bibitem[{Fisch et~al.(2019)Fisch, Talmor, Jia, Seo, Choi, and
  Chen}]{fisch-etal-2019-mrqa}
Adam Fisch, Alon Talmor, Robin Jia, Minjoon Seo, Eunsol Choi, and Danqi Chen.
  2019.
\newblock \href {https://doi.org/10.18653/v1/D19-5801} {{MRQA} 2019 shared
  task: Evaluating generalization in reading comprehension}.
\newblock In \emph{Proceedings of the 2nd Workshop on Machine Reading for
  Question Answering}, pages 1--13, Hong Kong, China. Association for
  Computational Linguistics.

\bibitem[{Ge et~al.(2024)Ge, Liu, Hu, Meng, Tao, Zhao, Ma, Zhang, Chen, Yang,
  Li, Xiao, and Zhu}]{ge2024clusteringrankingdiversitypreservedinstruction}
Yuan Ge, Yilun Liu, Chi Hu, Weibin Meng, Shimin Tao, Xiaofeng Zhao, Hongxia Ma,
  Li~Zhang, Boxing Chen, Hao Yang, Bei Li, Tong Xiao, and Jingbo Zhu. 2024.
\newblock \href {https://arxiv.org/abs/2402.18191} {Clustering and ranking:
  Diversity-preserved instruction selection through expert-aligned quality
  estimation}.
\newblock \emph{Preprint}, arXiv:2402.18191.

\bibitem[{Grattafiori et~al.(2024)Grattafiori, Dubey, Jauhri, Pandey, Kadian,
  Al-Dahle, Letman, Mathur, Schelten, Vaughan, Yang, Fan, Goyal, Hartshorn,
  Yang, Mitra, Sravankumar, Korenev, Hinsvark, Rao, Zhang, Rodriguez,
  Gregerson, Spataru, Roziere, Biron, Tang, Chern, Caucheteux, Nayak, Bi,
  Marra, McConnell, Keller, Touret, Wu, Wong, Ferrer, Nikolaidis, Allonsius,
  Song, Pintz, Livshits, Wyatt, Esiobu, Choudhary, Mahajan, Garcia-Olano,
  Perino, Hupkes, Lakomkin, AlBadawy, Lobanova, Dinan, Smith, Radenovic,
  Guzmán, Zhang, Synnaeve, Lee, Anderson, Thattai, Nail, Mialon, Pang,
  Cucurell, Nguyen, Korevaar, Xu, Touvron, Zarov, Ibarra, Kloumann, Misra,
  Evtimov, Zhang, Copet, Lee, Geffert, Vranes, Park, Mahadeokar, Shah, van~der
  Linde, Billock, Hong, Lee, Fu, Chi, Huang, Liu, Wang, Yu, Bitton, Spisak,
  Park, Rocca, Johnstun, Saxe, Jia, Alwala, Prasad, Upasani, Plawiak, Li,
  Heafield, Stone, El-Arini, Iyer, Malik, Chiu, Bhalla, Lakhotia,
  Rantala-Yeary, van~der Maaten, Chen, Tan, Jenkins, Martin, Madaan, Malo,
  Blecher, Landzaat, de~Oliveira, Muzzi, Pasupuleti, Singh, Paluri, Kardas,
  Tsimpoukelli, Oldham, Rita, Pavlova, Kambadur, Lewis, Si, Singh, Hassan,
  Goyal, Torabi, Bashlykov, Bogoychev, Chatterji, Zhang, Duchenne, Çelebi,
  Alrassy, Zhang, Li, Vasic, Weng, Bhargava, Dubal, Krishnan, Koura, Xu, He,
  Dong, Srinivasan, Ganapathy, Calderer, Cabral, Stojnic, Raileanu, Maheswari,
  Girdhar, Patel, Sauvestre, Polidoro, Sumbaly, Taylor, Silva, Hou, Wang,
  Hosseini, Chennabasappa, Singh, Bell, Kim, Edunov, Nie, Narang, Raparthy,
  Shen, Wan, Bhosale, Zhang, Vandenhende, Batra, Whitman, Sootla, Collot,
  Gururangan, Borodinsky, Herman, Fowler, Sheasha, Georgiou, Scialom,
  Speckbacher, Mihaylov, Xiao, Karn, Goswami, Gupta, Ramanathan, Kerkez,
  Gonguet, Do, Vogeti, Albiero, Petrovic, Chu, Xiong, Fu, Meers, Martinet,
  Wang, Wang, Tan, Xia, Xie, Jia, Wang, Goldschlag, Gaur, Babaei, Wen, Song,
  Zhang, Li, Mao, Coudert, Yan, Chen, Papakipos, Singh, Srivastava, Jain,
  Kelsey, Shajnfeld, Gangidi, Victoria, Goldstand, Menon, Sharma, Boesenberg,
  Baevski, Feinstein, Kallet, Sangani, Teo, Yunus, Lupu, Alvarado, Caples, Gu,
  Ho, Poulton, Ryan, Ramchandani, Dong, Franco, Goyal, Saraf, Chowdhury,
  Gabriel, Bharambe, Eisenman, Yazdan, James, Maurer, Leonhardi, Huang, Loyd,
  Paola, Paranjape, Liu, Wu, Ni, Hancock, Wasti, Spence, Stojkovic, Gamido,
  Montalvo, Parker, Burton, Mejia, Liu, Wang, Kim, Zhou, Hu, Chu, Cai, Tindal,
  Feichtenhofer, Gao, Civin, Beaty, Kreymer, Li, Adkins, Xu, Testuggine, David,
  Parikh, Liskovich, Foss, Wang, Le, Holland, Dowling, Jamil, Montgomery,
  Presani, Hahn, Wood, Le, Brinkman, Arcaute, Dunbar, Smothers, Sun, Kreuk,
  Tian, Kokkinos, Ozgenel, Caggioni, Kanayet, Seide, Florez, Schwarz, Badeer,
  Swee, Halpern, Herman, Sizov, Guangyi, Zhang, Lakshminarayanan, Inan,
  Shojanazeri, Zou, Wang, Zha, Habeeb, Rudolph, Suk, Aspegren, Goldman, Zhan,
  Damlaj, Molybog, Tufanov, Leontiadis, Veliche, Gat, Weissman, Geboski, Kohli,
  Lam, Asher, Gaya, Marcus, Tang, Chan, Zhen, Reizenstein, Teboul, Zhong, Jin,
  Yang, Cummings, Carvill, Shepard, McPhie, Torres, Ginsburg, Wang, Wu, U,
  Saxena, Khandelwal, Zand, Matosich, Veeraraghavan, Michelena, Li, Jagadeesh,
  Huang, Chawla, Huang, Chen, Garg, A, Silva, Bell, Zhang, Guo, Yu, Moshkovich,
  Wehrstedt, Khabsa, Avalani, Bhatt, Mankus, Hasson, Lennie, Reso, Groshev,
  Naumov, Lathi, Keneally, Liu, Seltzer, Valko, Restrepo, Patel, Vyatskov,
  Samvelyan, Clark, Macey, Wang, Hermoso, Metanat, Rastegari, Bansal,
  Santhanam, Parks, White, Bawa, Singhal, Egebo, Usunier, Mehta, Laptev, Dong,
  Cheng, Chernoguz, Hart, Salpekar, Kalinli, Kent, Parekh, Saab, Balaji,
  Rittner, Bontrager, Roux, Dollar, Zvyagina, Ratanchandani, Yuvraj, Liang,
  Alao, Rodriguez, Ayub, Murthy, Nayani, Mitra, Parthasarathy, Li, Hogan,
  Battey, Wang, Howes, Rinott, Mehta, Siby, Bondu, Datta, Chugh, Hunt, Dhillon,
  Sidorov, Pan, Mahajan, Verma, Yamamoto, Ramaswamy, Lindsay, Lindsay, Feng,
  Lin, Zha, Patil, Shankar, Zhang, Zhang, Wang, Agarwal, Sajuyigbe, Chintala,
  Max, Chen, Kehoe, Satterfield, Govindaprasad, Gupta, Deng, Cho, Virk,
  Subramanian, Choudhury, Goldman, Remez, Glaser, Best, Koehler, Robinson, Li,
  Zhang, Matthews, Chou, Shaked, Vontimitta, Ajayi, Montanez, Mohan, Kumar,
  Mangla, Ionescu, Poenaru, Mihailescu, Ivanov, Li, Wang, Jiang, Bouaziz,
  Constable, Tang, Wu, Wang, Wu, Gao, Kleinman, Chen, Hu, Jia, Qi, Li, Zhang,
  Zhang, Adi, Nam, Yu, Wang, Zhao, Hao, Qian, Li, He, Rait, DeVito, Rosnbrick,
  Wen, Yang, Zhao, and Ma}]{grattafiori2024llama3herdmodels}
Aaron Grattafiori, Abhimanyu Dubey, Abhinav Jauhri, Abhinav Pandey, Abhishek
  Kadian, Ahmad Al-Dahle, Aiesha Letman, Akhil Mathur, Alan Schelten, Alex
  Vaughan, Amy Yang, Angela Fan, Anirudh Goyal, Anthony Hartshorn, Aobo Yang,
  Archi Mitra, Archie Sravankumar, Artem Korenev, Arthur Hinsvark, Arun Rao,
  Aston Zhang, Aurelien Rodriguez, Austen Gregerson, Ava Spataru, Baptiste
  Roziere, Bethany Biron, Binh Tang, Bobbie Chern, Charlotte Caucheteux, Chaya
  Nayak, Chloe Bi, Chris Marra, Chris McConnell, Christian Keller, Christophe
  Touret, Chunyang Wu, Corinne Wong, Cristian~Canton Ferrer, Cyrus Nikolaidis,
  Damien Allonsius, Daniel Song, Danielle Pintz, Danny Livshits, Danny Wyatt,
  David Esiobu, Dhruv Choudhary, Dhruv Mahajan, Diego Garcia-Olano, Diego
  Perino, Dieuwke Hupkes, Egor Lakomkin, Ehab AlBadawy, Elina Lobanova, Emily
  Dinan, Eric~Michael Smith, Filip Radenovic, Francisco Guzmán, Frank Zhang,
  Gabriel Synnaeve, Gabrielle Lee, Georgia~Lewis Anderson, Govind Thattai,
  Graeme Nail, Gregoire Mialon, Guan Pang, Guillem Cucurell, Hailey Nguyen,
  Hannah Korevaar, Hu~Xu, Hugo Touvron, Iliyan Zarov, Imanol~Arrieta Ibarra,
  Isabel Kloumann, Ishan Misra, Ivan Evtimov, Jack Zhang, Jade Copet, Jaewon
  Lee, Jan Geffert, Jana Vranes, Jason Park, Jay Mahadeokar, Jeet Shah, Jelmer
  van~der Linde, Jennifer Billock, Jenny Hong, Jenya Lee, Jeremy Fu, Jianfeng
  Chi, Jianyu Huang, Jiawen Liu, Jie Wang, Jiecao Yu, Joanna Bitton, Joe
  Spisak, Jongsoo Park, Joseph Rocca, Joshua Johnstun, Joshua Saxe, Junteng
  Jia, Kalyan~Vasuden Alwala, Karthik Prasad, Kartikeya Upasani, Kate Plawiak,
  Ke~Li, Kenneth Heafield, Kevin Stone, Khalid El-Arini, Krithika Iyer, Kshitiz
  Malik, Kuenley Chiu, Kunal Bhalla, Kushal Lakhotia, Lauren Rantala-Yeary,
  Laurens van~der Maaten, Lawrence Chen, Liang Tan, Liz Jenkins, Louis Martin,
  Lovish Madaan, Lubo Malo, Lukas Blecher, Lukas Landzaat, Luke de~Oliveira,
  Madeline Muzzi, Mahesh Pasupuleti, Mannat Singh, Manohar Paluri, Marcin
  Kardas, Maria Tsimpoukelli, Mathew Oldham, Mathieu Rita, Maya Pavlova,
  Melanie Kambadur, Mike Lewis, Min Si, Mitesh~Kumar Singh, Mona Hassan, Naman
  Goyal, Narjes Torabi, Nikolay Bashlykov, Nikolay Bogoychev, Niladri
  Chatterji, Ning Zhang, Olivier Duchenne, Onur Çelebi, Patrick Alrassy,
  Pengchuan Zhang, Pengwei Li, Petar Vasic, Peter Weng, Prajjwal Bhargava,
  Pratik Dubal, Praveen Krishnan, Punit~Singh Koura, Puxin Xu, Qing He,
  Qingxiao Dong, Ragavan Srinivasan, Raj Ganapathy, Ramon Calderer,
  Ricardo~Silveira Cabral, Robert Stojnic, Roberta Raileanu, Rohan Maheswari,
  Rohit Girdhar, Rohit Patel, Romain Sauvestre, Ronnie Polidoro, Roshan
  Sumbaly, Ross Taylor, Ruan Silva, Rui Hou, Rui Wang, Saghar Hosseini, Sahana
  Chennabasappa, Sanjay Singh, Sean Bell, Seohyun~Sonia Kim, Sergey Edunov,
  Shaoliang Nie, Sharan Narang, Sharath Raparthy, Sheng Shen, Shengye Wan,
  Shruti Bhosale, Shun Zhang, Simon Vandenhende, Soumya Batra, Spencer Whitman,
  Sten Sootla, Stephane Collot, Suchin Gururangan, Sydney Borodinsky, Tamar
  Herman, Tara Fowler, Tarek Sheasha, Thomas Georgiou, Thomas Scialom, Tobias
  Speckbacher, Todor Mihaylov, Tong Xiao, Ujjwal Karn, Vedanuj Goswami, Vibhor
  Gupta, Vignesh Ramanathan, Viktor Kerkez, Vincent Gonguet, Virginie Do, Vish
  Vogeti, Vítor Albiero, Vladan Petrovic, Weiwei Chu, Wenhan Xiong, Wenyin Fu,
  Whitney Meers, Xavier Martinet, Xiaodong Wang, Xiaofang Wang, Xiaoqing~Ellen
  Tan, Xide Xia, Xinfeng Xie, Xuchao Jia, Xuewei Wang, Yaelle Goldschlag,
  Yashesh Gaur, Yasmine Babaei, Yi~Wen, Yiwen Song, Yuchen Zhang, Yue Li,
  Yuning Mao, Zacharie~Delpierre Coudert, Zheng Yan, Zhengxing Chen, Zoe
  Papakipos, Aaditya Singh, Aayushi Srivastava, Abha Jain, Adam Kelsey, Adam
  Shajnfeld, Adithya Gangidi, Adolfo Victoria, Ahuva Goldstand, Ajay Menon,
  Ajay Sharma, Alex Boesenberg, Alexei Baevski, Allie Feinstein, Amanda Kallet,
  Amit Sangani, Amos Teo, Anam Yunus, Andrei Lupu, Andres Alvarado, Andrew
  Caples, Andrew Gu, Andrew Ho, Andrew Poulton, Andrew Ryan, Ankit Ramchandani,
  Annie Dong, Annie Franco, Anuj Goyal, Aparajita Saraf, Arkabandhu Chowdhury,
  Ashley Gabriel, Ashwin Bharambe, Assaf Eisenman, Azadeh Yazdan, Beau James,
  Ben Maurer, Benjamin Leonhardi, Bernie Huang, Beth Loyd, Beto~De Paola,
  Bhargavi Paranjape, Bing Liu, Bo~Wu, Boyu Ni, Braden Hancock, Bram Wasti,
  Brandon Spence, Brani Stojkovic, Brian Gamido, Britt Montalvo, Carl Parker,
  Carly Burton, Catalina Mejia, Ce~Liu, Changhan Wang, Changkyu Kim, Chao Zhou,
  Chester Hu, Ching-Hsiang Chu, Chris Cai, Chris Tindal, Christoph
  Feichtenhofer, Cynthia Gao, Damon Civin, Dana Beaty, Daniel Kreymer, Daniel
  Li, David Adkins, David Xu, Davide Testuggine, Delia David, Devi Parikh,
  Diana Liskovich, Didem Foss, Dingkang Wang, Duc Le, Dustin Holland, Edward
  Dowling, Eissa Jamil, Elaine Montgomery, Eleonora Presani, Emily Hahn, Emily
  Wood, Eric-Tuan Le, Erik Brinkman, Esteban Arcaute, Evan Dunbar, Evan
  Smothers, Fei Sun, Felix Kreuk, Feng Tian, Filippos Kokkinos, Firat Ozgenel,
  Francesco Caggioni, Frank Kanayet, Frank Seide, Gabriela~Medina Florez,
  Gabriella Schwarz, Gada Badeer, Georgia Swee, Gil Halpern, Grant Herman,
  Grigory Sizov, Guangyi, Zhang, Guna Lakshminarayanan, Hakan Inan, Hamid
  Shojanazeri, Han Zou, Hannah Wang, Hanwen Zha, Haroun Habeeb, Harrison
  Rudolph, Helen Suk, Henry Aspegren, Hunter Goldman, Hongyuan Zhan, Ibrahim
  Damlaj, Igor Molybog, Igor Tufanov, Ilias Leontiadis, Irina-Elena Veliche,
  Itai Gat, Jake Weissman, James Geboski, James Kohli, Janice Lam, Japhet
  Asher, Jean-Baptiste Gaya, Jeff Marcus, Jeff Tang, Jennifer Chan, Jenny Zhen,
  Jeremy Reizenstein, Jeremy Teboul, Jessica Zhong, Jian Jin, Jingyi Yang, Joe
  Cummings, Jon Carvill, Jon Shepard, Jonathan McPhie, Jonathan Torres, Josh
  Ginsburg, Junjie Wang, Kai Wu, Kam~Hou U, Karan Saxena, Kartikay Khandelwal,
  Katayoun Zand, Kathy Matosich, Kaushik Veeraraghavan, Kelly Michelena, Keqian
  Li, Kiran Jagadeesh, Kun Huang, Kunal Chawla, Kyle Huang, Lailin Chen,
  Lakshya Garg, Lavender A, Leandro Silva, Lee Bell, Lei Zhang, Liangpeng Guo,
  Licheng Yu, Liron Moshkovich, Luca Wehrstedt, Madian Khabsa, Manav Avalani,
  Manish Bhatt, Martynas Mankus, Matan Hasson, Matthew Lennie, Matthias Reso,
  Maxim Groshev, Maxim Naumov, Maya Lathi, Meghan Keneally, Miao Liu,
  Michael~L. Seltzer, Michal Valko, Michelle Restrepo, Mihir Patel, Mik
  Vyatskov, Mikayel Samvelyan, Mike Clark, Mike Macey, Mike Wang, Miquel~Jubert
  Hermoso, Mo~Metanat, Mohammad Rastegari, Munish Bansal, Nandhini Santhanam,
  Natascha Parks, Natasha White, Navyata Bawa, Nayan Singhal, Nick Egebo,
  Nicolas Usunier, Nikhil Mehta, Nikolay~Pavlovich Laptev, Ning Dong, Norman
  Cheng, Oleg Chernoguz, Olivia Hart, Omkar Salpekar, Ozlem Kalinli, Parkin
  Kent, Parth Parekh, Paul Saab, Pavan Balaji, Pedro Rittner, Philip Bontrager,
  Pierre Roux, Piotr Dollar, Polina Zvyagina, Prashant Ratanchandani, Pritish
  Yuvraj, Qian Liang, Rachad Alao, Rachel Rodriguez, Rafi Ayub, Raghotham
  Murthy, Raghu Nayani, Rahul Mitra, Rangaprabhu Parthasarathy, Raymond Li,
  Rebekkah Hogan, Robin Battey, Rocky Wang, Russ Howes, Ruty Rinott, Sachin
  Mehta, Sachin Siby, Sai~Jayesh Bondu, Samyak Datta, Sara Chugh, Sara Hunt,
  Sargun Dhillon, Sasha Sidorov, Satadru Pan, Saurabh Mahajan, Saurabh Verma,
  Seiji Yamamoto, Sharadh Ramaswamy, Shaun Lindsay, Shaun Lindsay, Sheng Feng,
  Shenghao Lin, Shengxin~Cindy Zha, Shishir Patil, Shiva Shankar, Shuqiang
  Zhang, Shuqiang Zhang, Sinong Wang, Sneha Agarwal, Soji Sajuyigbe, Soumith
  Chintala, Stephanie Max, Stephen Chen, Steve Kehoe, Steve Satterfield,
  Sudarshan Govindaprasad, Sumit Gupta, Summer Deng, Sungmin Cho, Sunny Virk,
  Suraj Subramanian, Sy~Choudhury, Sydney Goldman, Tal Remez, Tamar Glaser,
  Tamara Best, Thilo Koehler, Thomas Robinson, Tianhe Li, Tianjun Zhang, Tim
  Matthews, Timothy Chou, Tzook Shaked, Varun Vontimitta, Victoria Ajayi,
  Victoria Montanez, Vijai Mohan, Vinay~Satish Kumar, Vishal Mangla, Vlad
  Ionescu, Vlad Poenaru, Vlad~Tiberiu Mihailescu, Vladimir Ivanov, Wei Li,
  Wenchen Wang, Wenwen Jiang, Wes Bouaziz, Will Constable, Xiaocheng Tang,
  Xiaojian Wu, Xiaolan Wang, Xilun Wu, Xinbo Gao, Yaniv Kleinman, Yanjun Chen,
  Ye~Hu, Ye~Jia, Ye~Qi, Yenda Li, Yilin Zhang, Ying Zhang, Yossi Adi, Youngjin
  Nam, Yu, Wang, Yu~Zhao, Yuchen Hao, Yundi Qian, Yunlu Li, Yuzi He, Zach Rait,
  Zachary DeVito, Zef Rosnbrick, Zhaoduo Wen, Zhenyu Yang, Zhiwei Zhao, and
  Zhiyu Ma. 2024.
\newblock \href {https://arxiv.org/abs/2407.21783} {The llama 3 herd of
  models}.
\newblock \emph{Preprint}, arXiv:2407.21783.

\bibitem[{Guo et~al.(2024)Guo, Yao, Shen, Wei, Zhang, Wang, and
  Liu}]{guo2024humaninstructionfreellmselfalignmentlimited}
Hongyi Guo, Yuanshun Yao, Wei Shen, Jiaheng Wei, Xiaoying Zhang, Zhaoran Wang,
  and Yang Liu. 2024.
\newblock \href {https://arxiv.org/abs/2401.06785} {Human-instruction-free llm
  self-alignment with limited samples}.
\newblock \emph{Preprint}, arXiv:2401.06785.

\bibitem[{Honovich et~al.(2023)Honovich, Scialom, Levy, and
  Schick}]{honovich-etal-2023-unnatural}
Or~Honovich, Thomas Scialom, Omer Levy, and Timo Schick. 2023.
\newblock \href {https://doi.org/10.18653/v1/2023.acl-long.806} {Unnatural
  instructions: Tuning language models with (almost) no human labor}.
\newblock In \emph{Proceedings of the 61st Annual Meeting of the Association
  for Computational Linguistics (Volume 1: Long Papers)}, pages 14409--14428,
  Toronto, Canada. Association for Computational Linguistics.

\bibitem[{Jiang et~al.(2024)Jiang, Sun, Shi, Rodriguez, Zhou, Neubig, Lin, Yih,
  and Iyer}]{jiang-etal-2024-instruction}
Zhengbao Jiang, Zhiqing Sun, Weijia Shi, Pedro Rodriguez, Chunting Zhou, Graham
  Neubig, Xi~Lin, Wen-tau Yih, and Srini Iyer. 2024.
\newblock \href {https://doi.org/10.18653/v1/2024.acl-long.296}
  {Instruction-tuned language models are better knowledge learners}.
\newblock In \emph{Proceedings of the 62nd Annual Meeting of the Association
  for Computational Linguistics (Volume 1: Long Papers)}, pages 5421--5434,
  Bangkok, Thailand. Association for Computational Linguistics.

\bibitem[{Köksal et~al.(2024)Köksal, Schick, Korhonen, and
  Schütze}]{köksal2024longformeffectiveinstructiontuning}
Abdullatif Köksal, Timo Schick, Anna Korhonen, and Hinrich Schütze. 2024.
\newblock \href {https://arxiv.org/abs/2304.08460} {Longform: Effective
  instruction tuning with reverse instructions}.
\newblock \emph{Preprint}, arXiv:2304.08460.

\bibitem[{Li et~al.(2024{\natexlab{a}})Li, Dong, Tang, Wang, Zhang, Huang,
  Huang, Huang, Huang, Zhang, Gu, Cheng, Wang, Chen, Dong, Lu, Sui, Wang, Lam,
  and Wei}]{li2024syntheticdataalmostscratch}
Haoran Li, Qingxiu Dong, Zhengyang Tang, Chaojun Wang, Xingxing Zhang, Haoyang
  Huang, Shaohan Huang, Xiaolong Huang, Zeqiang Huang, Dongdong Zhang, Yuxian
  Gu, Xin Cheng, Xun Wang, Si-Qing Chen, Li~Dong, Wei Lu, Zhifang Sui, Benyou
  Wang, Wai Lam, and Furu Wei. 2024{\natexlab{a}}.
\newblock \href {https://arxiv.org/abs/2402.13064} {Synthetic data (almost)
  from scratch: Generalized instruction tuning for language models}.
\newblock \emph{Preprint}, arXiv:2402.13064.

\bibitem[{Li et~al.(2024{\natexlab{b}})Li, Yu, Zhou, Schick, Levy, Zettlemoyer,
  Weston, and Lewis}]{li2024selfalignmentinstructionbacktranslation}
Xian Li, Ping Yu, Chunting Zhou, Timo Schick, Omer Levy, Luke Zettlemoyer,
  Jason Weston, and Mike Lewis. 2024{\natexlab{b}}.
\newblock \href {https://arxiv.org/abs/2308.06259} {Self-alignment with
  instruction backtranslation}.
\newblock \emph{Preprint}, arXiv:2308.06259.

\bibitem[{Liu et~al.(2024)Liu, Wei, Liu, Si, Zhang, Rao, Zheng, Peng, Yang,
  Zhou, and Dai}]{liu2024bestpracticeslessonslearned}
Ruibo Liu, Jerry Wei, Fangyu Liu, Chenglei Si, Yanzhe Zhang, Jinmeng Rao,
  Steven Zheng, Daiyi Peng, Diyi Yang, Denny Zhou, and Andrew~M. Dai. 2024.
\newblock \href {https://arxiv.org/abs/2404.07503} {Best practices and lessons
  learned on synthetic data}.
\newblock \emph{Preprint}, arXiv:2404.07503.

\bibitem[{Long et~al.(2024)Long, Wang, Xiao, Zhao, Ding, Chen, and
  Wang}]{long2024llmsdrivensyntheticdatageneration}
Lin Long, Rui Wang, Ruixuan Xiao, Junbo Zhao, Xiao Ding, Gang Chen, and Haobo
  Wang. 2024.
\newblock \href {https://arxiv.org/abs/2406.15126} {On llms-driven synthetic
  data generation, curation, and evaluation: A survey}.
\newblock \emph{Preprint}, arXiv:2406.15126.

\bibitem[{Longpre et~al.(2023)Longpre, Hou, Vu, Webson, Chung, Tay, Zhou, Le,
  Zoph, Wei, and Roberts}]{longpre2023flancollectiondesigningdata}
Shayne Longpre, Le~Hou, Tu~Vu, Albert Webson, Hyung~Won Chung, Yi~Tay, Denny
  Zhou, Quoc~V. Le, Barret Zoph, Jason Wei, and Adam Roberts. 2023.
\newblock \href {https://arxiv.org/abs/2301.13688} {The flan collection:
  Designing data and methods for effective instruction tuning}.
\newblock \emph{Preprint}, arXiv:2301.13688.

\bibitem[{Mishra et~al.(2022)Mishra, Khashabi, Baral, and
  Hajishirzi}]{mishra-etal-2022-cross}
Swaroop Mishra, Daniel Khashabi, Chitta Baral, and Hannaneh Hajishirzi. 2022.
\newblock \href {https://doi.org/10.18653/v1/2022.acl-long.244} {Cross-task
  generalization via natural language crowdsourcing instructions}.
\newblock In \emph{Proceedings of the 60th Annual Meeting of the Association
  for Computational Linguistics (Volume 1: Long Papers)}, pages 3470--3487,
  Dublin, Ireland. Association for Computational Linguistics.

\bibitem[{Nayak et~al.(2024{\natexlab{a}})Nayak, Nan, Trost, and
  Bach}]{nayak-etal-2024-learning}
Nihal Nayak, Yiyang Nan, Avi Trost, and Stephen Bach. 2024{\natexlab{a}}.
\newblock \href {https://doi.org/10.18653/v1/2024.findings-acl.748} {Learning
  to generate instruction tuning datasets for zero-shot task adaptation}.
\newblock In \emph{Findings of the Association for Computational Linguistics
  ACL 2024}, pages 12585--12611, Bangkok, Thailand and virtual meeting.
  Association for Computational Linguistics.

\bibitem[{Nayak et~al.(2024{\natexlab{b}})Nayak, Nan, Trost, and
  Bach}]{nayak2024learninggenerateinstructiontuning}
Nihal~V. Nayak, Yiyang Nan, Avi Trost, and Stephen~H. Bach. 2024{\natexlab{b}}.
\newblock \href {https://arxiv.org/abs/2402.18334} {Learning to generate
  instruction tuning datasets for zero-shot task adaptation}.
\newblock \emph{Preprint}, arXiv:2402.18334.

\bibitem[{OpenAI et~al.(2024)OpenAI, Achiam, Adler, Agarwal, Ahmad, Akkaya,
  Aleman, Almeida, Altenschmidt, Altman, Anadkat, Avila, Babuschkin, Balaji,
  Balcom, Baltescu, Bao, Bavarian, Belgum, Bello, Berdine, Bernadett-Shapiro,
  Berner, Bogdonoff, Boiko, Boyd, Brakman, Brockman, Brooks, Brundage, Button,
  Cai, Campbell, Cann, Carey, Carlson, Carmichael, Chan, Chang, Chantzis, Chen,
  Chen, Chen, Chen, Chen, Chess, Cho, Chu, Chung, Cummings, Currier, Dai,
  Decareaux, Degry, Deutsch, Deville, Dhar, Dohan, Dowling, Dunning, Ecoffet,
  Eleti, Eloundou, Farhi, Fedus, Felix, Fishman, Forte, Fulford, Gao, Georges,
  Gibson, Goel, Gogineni, Goh, Gontijo-Lopes, Gordon, Grafstein, Gray, Greene,
  Gross, Gu, Guo, Hallacy, Han, Harris, He, Heaton, Heidecke, Hesse, Hickey,
  Hickey, Hoeschele, Houghton, Hsu, Hu, Hu, Huizinga, Jain, Jain, Jang, Jiang,
  Jiang, Jin, Jin, Jomoto, Jonn, Jun, Kaftan, Łukasz Kaiser, Kamali,
  Kanitscheider, Keskar, Khan, Kilpatrick, Kim, Kim, Kim, Kirchner, Kiros,
  Knight, Kokotajlo, Łukasz Kondraciuk, Kondrich, Konstantinidis, Kosic,
  Krueger, Kuo, Lampe, Lan, Lee, Leike, Leung, Levy, Li, Lim, Lin, Lin, Litwin,
  Lopez, Lowe, Lue, Makanju, Malfacini, Manning, Markov, Markovski, Martin,
  Mayer, Mayne, McGrew, McKinney, McLeavey, McMillan, McNeil, Medina, Mehta,
  Menick, Metz, Mishchenko, Mishkin, Monaco, Morikawa, Mossing, Mu, Murati,
  Murk, Mély, Nair, Nakano, Nayak, Neelakantan, Ngo, Noh, Ouyang, O'Keefe,
  Pachocki, Paino, Palermo, Pantuliano, Parascandolo, Parish, Parparita,
  Passos, Pavlov, Peng, Perelman, de~Avila Belbute~Peres, Petrov,
  de~Oliveira~Pinto, Michael, Pokorny, Pokrass, Pong, Powell, Power, Power,
  Proehl, Puri, Radford, Rae, Ramesh, Raymond, Real, Rimbach, Ross, Rotsted,
  Roussez, Ryder, Saltarelli, Sanders, Santurkar, Sastry, Schmidt, Schnurr,
  Schulman, Selsam, Sheppard, Sherbakov, Shieh, Shoker, Shyam, Sidor, Sigler,
  Simens, Sitkin, Slama, Sohl, Sokolowsky, Song, Staudacher, Such, Summers,
  Sutskever, Tang, Tezak, Thompson, Tillet, Tootoonchian, Tseng, Tuggle,
  Turley, Tworek, Uribe, Vallone, Vijayvergiya, Voss, Wainwright, Wang, Wang,
  Wang, Ward, Wei, Weinmann, Welihinda, Welinder, Weng, Weng, Wiethoff,
  Willner, Winter, Wolrich, Wong, Workman, Wu, Wu, Wu, Xiao, Xu, Yoo, Yu, Yuan,
  Zaremba, Zellers, Zhang, Zhang, Zhao, Zheng, Zhuang, Zhuk, and
  Zoph}]{openai2024gpt4technicalreport}
OpenAI, Josh Achiam, Steven Adler, Sandhini Agarwal, Lama Ahmad, Ilge Akkaya,
  Florencia~Leoni Aleman, Diogo Almeida, Janko Altenschmidt, Sam Altman,
  Shyamal Anadkat, Red Avila, Igor Babuschkin, Suchir Balaji, Valerie Balcom,
  Paul Baltescu, Haiming Bao, Mohammad Bavarian, Jeff Belgum, Irwan Bello, Jake
  Berdine, Gabriel Bernadett-Shapiro, Christopher Berner, Lenny Bogdonoff, Oleg
  Boiko, Madelaine Boyd, Anna-Luisa Brakman, Greg Brockman, Tim Brooks, Miles
  Brundage, Kevin Button, Trevor Cai, Rosie Campbell, Andrew Cann, Brittany
  Carey, Chelsea Carlson, Rory Carmichael, Brooke Chan, Che Chang, Fotis
  Chantzis, Derek Chen, Sully Chen, Ruby Chen, Jason Chen, Mark Chen, Ben
  Chess, Chester Cho, Casey Chu, Hyung~Won Chung, Dave Cummings, Jeremiah
  Currier, Yunxing Dai, Cory Decareaux, Thomas Degry, Noah Deutsch, Damien
  Deville, Arka Dhar, David Dohan, Steve Dowling, Sheila Dunning, Adrien
  Ecoffet, Atty Eleti, Tyna Eloundou, David Farhi, Liam Fedus, Niko Felix,
  Simón~Posada Fishman, Juston Forte, Isabella Fulford, Leo Gao, Elie Georges,
  Christian Gibson, Vik Goel, Tarun Gogineni, Gabriel Goh, Rapha Gontijo-Lopes,
  Jonathan Gordon, Morgan Grafstein, Scott Gray, Ryan Greene, Joshua Gross,
  Shixiang~Shane Gu, Yufei Guo, Chris Hallacy, Jesse Han, Jeff Harris, Yuchen
  He, Mike Heaton, Johannes Heidecke, Chris Hesse, Alan Hickey, Wade Hickey,
  Peter Hoeschele, Brandon Houghton, Kenny Hsu, Shengli Hu, Xin Hu, Joost
  Huizinga, Shantanu Jain, Shawn Jain, Joanne Jang, Angela Jiang, Roger Jiang,
  Haozhun Jin, Denny Jin, Shino Jomoto, Billie Jonn, Heewoo Jun, Tomer Kaftan,
  Łukasz Kaiser, Ali Kamali, Ingmar Kanitscheider, Nitish~Shirish Keskar,
  Tabarak Khan, Logan Kilpatrick, Jong~Wook Kim, Christina Kim, Yongjik Kim,
  Jan~Hendrik Kirchner, Jamie Kiros, Matt Knight, Daniel Kokotajlo, Łukasz
  Kondraciuk, Andrew Kondrich, Aris Konstantinidis, Kyle Kosic, Gretchen
  Krueger, Vishal Kuo, Michael Lampe, Ikai Lan, Teddy Lee, Jan Leike, Jade
  Leung, Daniel Levy, Chak~Ming Li, Rachel Lim, Molly Lin, Stephanie Lin,
  Mateusz Litwin, Theresa Lopez, Ryan Lowe, Patricia Lue, Anna Makanju, Kim
  Malfacini, Sam Manning, Todor Markov, Yaniv Markovski, Bianca Martin, Katie
  Mayer, Andrew Mayne, Bob McGrew, Scott~Mayer McKinney, Christine McLeavey,
  Paul McMillan, Jake McNeil, David Medina, Aalok Mehta, Jacob Menick, Luke
  Metz, Andrey Mishchenko, Pamela Mishkin, Vinnie Monaco, Evan Morikawa, Daniel
  Mossing, Tong Mu, Mira Murati, Oleg Murk, David Mély, Ashvin Nair, Reiichiro
  Nakano, Rajeev Nayak, Arvind Neelakantan, Richard Ngo, Hyeonwoo Noh, Long
  Ouyang, Cullen O'Keefe, Jakub Pachocki, Alex Paino, Joe Palermo, Ashley
  Pantuliano, Giambattista Parascandolo, Joel Parish, Emy Parparita, Alex
  Passos, Mikhail Pavlov, Andrew Peng, Adam Perelman, Filipe de~Avila
  Belbute~Peres, Michael Petrov, Henrique~Ponde de~Oliveira~Pinto, Michael,
  Pokorny, Michelle Pokrass, Vitchyr~H. Pong, Tolly Powell, Alethea Power,
  Boris Power, Elizabeth Proehl, Raul Puri, Alec Radford, Jack Rae, Aditya
  Ramesh, Cameron Raymond, Francis Real, Kendra Rimbach, Carl Ross, Bob
  Rotsted, Henri Roussez, Nick Ryder, Mario Saltarelli, Ted Sanders, Shibani
  Santurkar, Girish Sastry, Heather Schmidt, David Schnurr, John Schulman,
  Daniel Selsam, Kyla Sheppard, Toki Sherbakov, Jessica Shieh, Sarah Shoker,
  Pranav Shyam, Szymon Sidor, Eric Sigler, Maddie Simens, Jordan Sitkin,
  Katarina Slama, Ian Sohl, Benjamin Sokolowsky, Yang Song, Natalie Staudacher,
  Felipe~Petroski Such, Natalie Summers, Ilya Sutskever, Jie Tang, Nikolas
  Tezak, Madeleine~B. Thompson, Phil Tillet, Amin Tootoonchian, Elizabeth
  Tseng, Preston Tuggle, Nick Turley, Jerry Tworek, Juan Felipe~Cerón Uribe,
  Andrea Vallone, Arun Vijayvergiya, Chelsea Voss, Carroll Wainwright,
  Justin~Jay Wang, Alvin Wang, Ben Wang, Jonathan Ward, Jason Wei, CJ~Weinmann,
  Akila Welihinda, Peter Welinder, Jiayi Weng, Lilian Weng, Matt Wiethoff, Dave
  Willner, Clemens Winter, Samuel Wolrich, Hannah Wong, Lauren Workman, Sherwin
  Wu, Jeff Wu, Michael Wu, Kai Xiao, Tao Xu, Sarah Yoo, Kevin Yu, Qiming Yuan,
  Wojciech Zaremba, Rowan Zellers, Chong Zhang, Marvin Zhang, Shengjia Zhao,
  Tianhao Zheng, Juntang Zhuang, William Zhuk, and Barret Zoph. 2024.
\newblock \href {https://arxiv.org/abs/2303.08774} {Gpt-4 technical report}.
\newblock \emph{Preprint}, arXiv:2303.08774.

\bibitem[{Peng et~al.(2023)Peng, Li, He, Galley, and
  Gao}]{peng2023instructiontuninggpt4}
Baolin Peng, Chunyuan Li, Pengcheng He, Michel Galley, and Jianfeng Gao. 2023.
\newblock \href {https://arxiv.org/abs/2304.03277} {Instruction tuning with
  gpt-4}.
\newblock \emph{Preprint}, arXiv:2304.03277.

\bibitem[{Perez et~al.(2022)Perez, Huang, Song, Cai, Ring, Aslanides, Glaese,
  McAleese, and Irving}]{perez2022redteaminglanguagemodels}
Ethan Perez, Saffron Huang, Francis Song, Trevor Cai, Roman Ring, John
  Aslanides, Amelia Glaese, Nat McAleese, and Geoffrey Irving. 2022.
\newblock \href {https://arxiv.org/abs/2202.03286} {Red teaming language models
  with language models}.
\newblock \emph{Preprint}, arXiv:2202.03286.

\bibitem[{Qwen et~al.(2025)Qwen, :, Yang, Yang, Zhang, Hui, Zheng, Yu, Li, Liu,
  Huang, Wei, Lin, Yang, Tu, Zhang, Yang, Yang, Zhou, Lin, Dang, Lu, Bao, Yang,
  Yu, Li, Xue, Zhang, Zhu, Men, Lin, Li, Tang, Xia, Ren, Ren, Fan, Su, Zhang,
  Wan, Liu, Cui, Zhang, and Qiu}]{qwen2.5}
Qwen, :, An~Yang, Baosong Yang, Beichen Zhang, Binyuan Hui, Bo~Zheng, Bowen Yu,
  Chengyuan Li, Dayiheng Liu, Fei Huang, Haoran Wei, Huan Lin, Jian Yang,
  Jianhong Tu, Jianwei Zhang, Jianxin Yang, Jiaxi Yang, Jingren Zhou, Junyang
  Lin, Kai Dang, Keming Lu, Keqin Bao, Kexin Yang, Le~Yu, Mei Li, Mingfeng Xue,
  Pei Zhang, Qin Zhu, Rui Men, Runji Lin, Tianhao Li, Tianyi Tang, Tingyu Xia,
  Xingzhang Ren, Xuancheng Ren, Yang Fan, Yang Su, Yichang Zhang, Yu~Wan,
  Yuqiong Liu, Zeyu Cui, Zhenru Zhang, and Zihan Qiu. 2025.
\newblock \href {https://arxiv.org/abs/2412.15115} {Qwen2.5 technical report}.
\newblock \emph{Preprint}, arXiv:2412.15115.

\bibitem[{Rajpurkar et~al.(2016)Rajpurkar, Zhang, Lopyrev, and
  Liang}]{rajpurkar-etal-2016-squad}
Pranav Rajpurkar, Jian Zhang, Konstantin Lopyrev, and Percy Liang. 2016.
\newblock \href {https://doi.org/10.18653/v1/D16-1264} {{SQ}u{AD}: 100,000+
  questions for machine comprehension of text}.
\newblock In \emph{Proceedings of the 2016 Conference on Empirical Methods in
  Natural Language Processing}, pages 2383--2392, Austin, Texas. Association
  for Computational Linguistics.

\bibitem[{Reimers and
  Gurevych(2019{\natexlab{a}})}]{reimers-gurevych-2019-sentence}
Nils Reimers and Iryna Gurevych. 2019{\natexlab{a}}.
\newblock \href {https://doi.org/10.18653/v1/D19-1410} {Sentence-{BERT}:
  Sentence embeddings using {S}iamese {BERT}-networks}.
\newblock In \emph{Proceedings of the 2019 Conference on Empirical Methods in
  Natural Language Processing and the 9th International Joint Conference on
  Natural Language Processing (EMNLP-IJCNLP)}, pages 3982--3992, Hong Kong,
  China. Association for Computational Linguistics.

\bibitem[{Reimers and
  Gurevych(2019{\natexlab{b}})}]{reimers-2019-sentence-bert}
Nils Reimers and Iryna Gurevych. 2019{\natexlab{b}}.
\newblock \href {http://arxiv.org/abs/1908.10084} {Sentence-bert: Sentence
  embeddings using siamese bert-networks}.
\newblock In \emph{Proceedings of the 2019 Conference on Empirical Methods in
  Natural Language Processing}. Association for Computational Linguistics.

\bibitem[{Saito et~al.(2024)Saito, Sohn, Lee, and Ushiku}]{wiki2023plus}
Kuniaki Saito, Kihyuk Sohn, Chen-Yu Lee, and Yoshitaka Ushiku. 2024.
\newblock \href {https://arxiv.org/abs/2402.12170} {Where is the answer?
  investigating positional bias in language model knowledge extraction}.
\newblock \emph{Preprint}, arXiv:2402.12170.

\bibitem[{Sanh et~al.(2022)Sanh, Webson, Raffel, Bach, Sutawika, Alyafeai,
  Chaffin, Stiegler, Scao, Raja, Dey, Bari, Xu, Thakker, Sharma, Szczechla,
  Kim, Chhablani, Nayak, Datta, Chang, Jiang, Wang, Manica, Shen, Yong, Pandey,
  Bawden, Wang, Neeraj, Rozen, Sharma, Santilli, Fevry, Fries, Teehan, Bers,
  Biderman, Gao, Wolf, and Rush}]{sanh2022multitaskpromptedtrainingenables}
Victor Sanh, Albert Webson, Colin Raffel, Stephen~H. Bach, Lintang Sutawika,
  Zaid Alyafeai, Antoine Chaffin, Arnaud Stiegler, Teven~Le Scao, Arun Raja,
  Manan Dey, M~Saiful Bari, Canwen Xu, Urmish Thakker, Shanya~Sharma Sharma,
  Eliza Szczechla, Taewoon Kim, Gunjan Chhablani, Nihal Nayak, Debajyoti Datta,
  Jonathan Chang, Mike Tian-Jian Jiang, Han Wang, Matteo Manica, Sheng Shen,
  Zheng~Xin Yong, Harshit Pandey, Rachel Bawden, Thomas Wang, Trishala Neeraj,
  Jos Rozen, Abheesht Sharma, Andrea Santilli, Thibault Fevry, Jason~Alan
  Fries, Ryan Teehan, Tali Bers, Stella Biderman, Leo Gao, Thomas Wolf, and
  Alexander~M. Rush. 2022.
\newblock \href {https://arxiv.org/abs/2110.08207} {Multitask prompted training
  enables zero-shot task generalization}.
\newblock \emph{Preprint}, arXiv:2110.08207.

\bibitem[{Schick and
  Schütze(2021)}]{schick2021generatingdatasetspretrainedlanguage}
Timo Schick and Hinrich Schütze. 2021.
\newblock \href {https://arxiv.org/abs/2104.07540} {Generating datasets with
  pretrained language models}.
\newblock \emph{Preprint}, arXiv:2104.07540.

\bibitem[{Sun et~al.(2023)Sun, Shen, Zhou, Zhang, Chen, Cox, Yang, and
  Gan}]{sun2023principledrivenselfalignmentlanguagemodels}
Zhiqing Sun, Yikang Shen, Qinhong Zhou, Hongxin Zhang, Zhenfang Chen, David
  Cox, Yiming Yang, and Chuang Gan. 2023.
\newblock \href {https://arxiv.org/abs/2305.03047} {Principle-driven
  self-alignment of language models from scratch with minimal human
  supervision}.
\newblock \emph{Preprint}, arXiv:2305.03047.

\bibitem[{Taori et~al.(2023)Taori, Gulrajani, Zhang, Dubois, Li, Guestrin,
  Liang, and Hashimoto}]{alpaca}
Rohan Taori, Ishaan Gulrajani, Tianyi Zhang, Yann Dubois, Xuechen Li, Carlos
  Guestrin, Percy Liang, and Tatsunori~B. Hashimoto. 2023.
\newblock Stanford alpaca: An instruction-following llama model.
\newblock \url{https://github.com/tatsu-lab/stanford_alpaca}.

\bibitem[{Team et~al.(2024)Team, Riviere, Pathak, Sessa, Hardin, Bhupatiraju,
  Hussenot, Mesnard, Shahriari, Ramé, Ferret, Liu, Tafti, Friesen, Casbon,
  Ramos, Kumar, Lan, Jerome, Tsitsulin, Vieillard, Stanczyk, Girgin, Momchev,
  Hoffman, Thakoor, Grill, Neyshabur, Bachem, Walton, Severyn, Parrish, Ahmad,
  Hutchison, Abdagic, Carl, Shen, Brock, Coenen, Laforge, Paterson, Bastian,
  Piot, Wu, Royal, Chen, Kumar, Perry, Welty, Choquette-Choo, Sinopalnikov,
  Weinberger, Vijaykumar, Rogozińska, Herbison, Bandy, Wang, Noland, Moreira,
  Senter, Eltyshev, Visin, Rasskin, Wei, Cameron, Martins, Hashemi,
  Klimczak-Plucińska, Batra, Dhand, Nardini, Mein, Zhou, Svensson, Stanway,
  Chan, Zhou, Carrasqueira, Iljazi, Becker, Fernandez, van Amersfoort, Gordon,
  Lipschultz, Newlan, yeong Ji, Mohamed, Badola, Black, Millican, McDonell,
  Nguyen, Sodhia, Greene, Sjoesund, Usui, Sifre, Heuermann, Lago, McNealus,
  Soares, Kilpatrick, Dixon, Martins, Reid, Singh, Iverson, Görner, Velloso,
  Wirth, Davidow, Miller, Rahtz, Watson, Risdal, Kazemi, Moynihan, Zhang,
  Kahng, Park, Rahman, Khatwani, Dao, Bardoliwalla, Devanathan, Dumai, Chauhan,
  Wahltinez, Botarda, Barnes, Barham, Michel, Jin, Georgiev, Culliton, Kuppala,
  Comanescu, Merhej, Jana, Rokni, Agarwal, Mullins, Saadat, Carthy, Cogan,
  Perrin, Arnold, Krause, Dai, Garg, Sheth, Ronstrom, Chan, Jordan, Yu, Eccles,
  Hennigan, Kocisky, Doshi, Jain, Yadav, Meshram, Dharmadhikari, Barkley, Wei,
  Ye, Han, Kwon, Xu, Shen, Gong, Wei, Cotruta, Kirk, Rao, Giang, Peran,
  Warkentin, Collins, Barral, Ghahramani, Hadsell, Sculley, Banks, Dragan,
  Petrov, Vinyals, Dean, Hassabis, Kavukcuoglu, Farabet, Buchatskaya, Borgeaud,
  Fiedel, Joulin, Kenealy, Dadashi, and
  Andreev}]{gemmateam2024gemma2improvingopen}
Gemma Team, Morgane Riviere, Shreya Pathak, Pier~Giuseppe Sessa, Cassidy
  Hardin, Surya Bhupatiraju, Léonard Hussenot, Thomas Mesnard, Bobak
  Shahriari, Alexandre Ramé, Johan Ferret, Peter Liu, Pouya Tafti, Abe
  Friesen, Michelle Casbon, Sabela Ramos, Ravin Kumar, Charline~Le Lan, Sammy
  Jerome, Anton Tsitsulin, Nino Vieillard, Piotr Stanczyk, Sertan Girgin,
  Nikola Momchev, Matt Hoffman, Shantanu Thakoor, Jean-Bastien Grill, Behnam
  Neyshabur, Olivier Bachem, Alanna Walton, Aliaksei Severyn, Alicia Parrish,
  Aliya Ahmad, Allen Hutchison, Alvin Abdagic, Amanda Carl, Amy Shen, Andy
  Brock, Andy Coenen, Anthony Laforge, Antonia Paterson, Ben Bastian, Bilal
  Piot, Bo~Wu, Brandon Royal, Charlie Chen, Chintu Kumar, Chris Perry, Chris
  Welty, Christopher~A. Choquette-Choo, Danila Sinopalnikov, David Weinberger,
  Dimple Vijaykumar, Dominika Rogozińska, Dustin Herbison, Elisa Bandy, Emma
  Wang, Eric Noland, Erica Moreira, Evan Senter, Evgenii Eltyshev, Francesco
  Visin, Gabriel Rasskin, Gary Wei, Glenn Cameron, Gus Martins, Hadi Hashemi,
  Hanna Klimczak-Plucińska, Harleen Batra, Harsh Dhand, Ivan Nardini, Jacinda
  Mein, Jack Zhou, James Svensson, Jeff Stanway, Jetha Chan, Jin~Peng Zhou,
  Joana Carrasqueira, Joana Iljazi, Jocelyn Becker, Joe Fernandez, Joost van
  Amersfoort, Josh Gordon, Josh Lipschultz, Josh Newlan, Ju~yeong Ji, Kareem
  Mohamed, Kartikeya Badola, Kat Black, Katie Millican, Keelin McDonell, Kelvin
  Nguyen, Kiranbir Sodhia, Kish Greene, Lars~Lowe Sjoesund, Lauren Usui,
  Laurent Sifre, Lena Heuermann, Leticia Lago, Lilly McNealus, Livio~Baldini
  Soares, Logan Kilpatrick, Lucas Dixon, Luciano Martins, Machel Reid,
  Manvinder Singh, Mark Iverson, Martin Görner, Mat Velloso, Mateo Wirth, Matt
  Davidow, Matt Miller, Matthew Rahtz, Matthew Watson, Meg Risdal, Mehran
  Kazemi, Michael Moynihan, Ming Zhang, Minsuk Kahng, Minwoo Park, Mofi Rahman,
  Mohit Khatwani, Natalie Dao, Nenshad Bardoliwalla, Nesh Devanathan, Neta
  Dumai, Nilay Chauhan, Oscar Wahltinez, Pankil Botarda, Parker Barnes, Paul
  Barham, Paul Michel, Pengchong Jin, Petko Georgiev, Phil Culliton, Pradeep
  Kuppala, Ramona Comanescu, Ramona Merhej, Reena Jana, Reza~Ardeshir Rokni,
  Rishabh Agarwal, Ryan Mullins, Samaneh Saadat, Sara~Mc Carthy, Sarah Cogan,
  Sarah Perrin, Sébastien M.~R. Arnold, Sebastian Krause, Shengyang Dai,
  Shruti Garg, Shruti Sheth, Sue Ronstrom, Susan Chan, Timothy Jordan, Ting Yu,
  Tom Eccles, Tom Hennigan, Tomas Kocisky, Tulsee Doshi, Vihan Jain, Vikas
  Yadav, Vilobh Meshram, Vishal Dharmadhikari, Warren Barkley, Wei Wei, Wenming
  Ye, Woohyun Han, Woosuk Kwon, Xiang Xu, Zhe Shen, Zhitao Gong, Zichuan Wei,
  Victor Cotruta, Phoebe Kirk, Anand Rao, Minh Giang, Ludovic Peran, Tris
  Warkentin, Eli Collins, Joelle Barral, Zoubin Ghahramani, Raia Hadsell,
  D.~Sculley, Jeanine Banks, Anca Dragan, Slav Petrov, Oriol Vinyals, Jeff
  Dean, Demis Hassabis, Koray Kavukcuoglu, Clement Farabet, Elena Buchatskaya,
  Sebastian Borgeaud, Noah Fiedel, Armand Joulin, Kathleen Kenealy, Robert
  Dadashi, and Alek Andreev. 2024.
\newblock \href {https://arxiv.org/abs/2408.00118} {Gemma 2: Improving open
  language models at a practical size}.
\newblock \emph{Preprint}, arXiv:2408.00118.

\bibitem[{Tevet and
  Berant(2021)}]{tevet2021evaluatingevaluationdiversitynatural}
Guy Tevet and Jonathan Berant. 2021.
\newblock \href {https://arxiv.org/abs/2004.02990} {Evaluating the evaluation
  of diversity in natural language generation}.
\newblock \emph{Preprint}, arXiv:2004.02990.

\bibitem[{Villalobos et~al.(2024)Villalobos, Ho, Sevilla, Besiroglu, Heim, and
  Hobbhahn}]{villalobos2024rundatalimitsllm}
Pablo Villalobos, Anson Ho, Jaime Sevilla, Tamay Besiroglu, Lennart Heim, and
  Marius Hobbhahn. 2024.
\newblock \href {https://arxiv.org/abs/2211.04325} {Will we run out of data?
  limits of llm scaling based on human-generated data}.
\newblock \emph{Preprint}, arXiv:2211.04325.

\bibitem[{Wandell(1995)}]{wandell1995foundations}
Brian~A. Wandell. 1995.
\newblock \emph{Foundations of Vision}.
\newblock Sinauer Associates, Sunderland, MA.

\bibitem[{Wang et~al.(2023{\natexlab{a}})Wang, Kordi, Mishra, Liu, Smith,
  Khashabi, and Hajishirzi}]{wang2023selfinstructaligninglanguagemodels}
Yizhong Wang, Yeganeh Kordi, Swaroop Mishra, Alisa Liu, Noah~A. Smith, Daniel
  Khashabi, and Hannaneh Hajishirzi. 2023{\natexlab{a}}.
\newblock \href {https://doi.org/10.18653/v1/2023.acl-long.754} {Self-instruct:
  Aligning language models with self-generated instructions}.
\newblock In \emph{Proceedings of the 61st Annual Meeting of the Association
  for Computational Linguistics (Volume 1: Long Papers)}, pages 13484--13508,
  Toronto, Canada. Association for Computational Linguistics.

\bibitem[{Wang et~al.(2023{\natexlab{b}})Wang, Zhong, Li, Mi, Zeng, Huang,
  Shang, Jiang, and Liu}]{wang2023aligninglargelanguagemodels}
Yufei Wang, Wanjun Zhong, Liangyou Li, Fei Mi, Xingshan Zeng, Wenyong Huang,
  Lifeng Shang, Xin Jiang, and Qun Liu. 2023{\natexlab{b}}.
\newblock \href {https://arxiv.org/abs/2307.12966} {Aligning large language
  models with human: A survey}.
\newblock \emph{Preprint}, arXiv:2307.12966.

\bibitem[{Wei et~al.(2022)Wei, Bosma, Zhao, Guu, Yu, Lester, Du, Dai, and
  Le}]{wei2022finetunedlanguagemodelszeroshot}
Jason Wei, Maarten Bosma, Vincent~Y. Zhao, Kelvin Guu, Adams~Wei Yu, Brian
  Lester, Nan Du, Andrew~M. Dai, and Quoc~V. Le. 2022.
\newblock \href {https://arxiv.org/abs/2109.01652} {Finetuned language models
  are zero-shot learners}.
\newblock \emph{Preprint}, arXiv:2109.01652.

\bibitem[{Wu et~al.(2024)Wu, Huang, Gao, Chen, Zhang, Wan, Zhou, Zhang, Gao,
  Xiao, and Sun}]{wu2024unigenunifiedframeworktextual}
Siyuan Wu, Yue Huang, Chujie Gao, Dongping Chen, Qihui Zhang, Yao Wan, Tianyi
  Zhou, Xiangliang Zhang, Jianfeng Gao, Chaowei Xiao, and Lichao Sun. 2024.
\newblock \href {https://arxiv.org/abs/2406.18966} {Unigen: A unified framework
  for textual dataset generation using large language models}.
\newblock \emph{Preprint}, arXiv:2406.18966.

\bibitem[{Xu et~al.(2023)Xu, Guo, Duan, and
  McAuley}]{xu2023baizeopensourcechatmodel}
Canwen Xu, Daya Guo, Nan Duan, and Julian McAuley. 2023.
\newblock \href {https://arxiv.org/abs/2304.01196} {Baize: An open-source chat
  model with parameter-efficient tuning on self-chat data}.
\newblock \emph{Preprint}, arXiv:2304.01196.

\bibitem[{Yang et~al.(2018)Yang, Qi, Zhang, Bengio, Cohen, Salakhutdinov, and
  Manning}]{yang-etal-2018-hotpotqa}
Zhilin Yang, Peng Qi, Saizheng Zhang, Yoshua Bengio, William Cohen, Ruslan
  Salakhutdinov, and Christopher~D. Manning. 2018.
\newblock \href {https://doi.org/10.18653/v1/D18-1259} {{H}otpot{QA}: A dataset
  for diverse, explainable multi-hop question answering}.
\newblock In \emph{Proceedings of the 2018 Conference on Empirical Methods in
  Natural Language Processing}, pages 2369--2380, Brussels, Belgium.
  Association for Computational Linguistics.

\bibitem[{Yuan et~al.(2024)Yuan, Pang, Cho, Li, Sukhbaatar, Xu, and
  Weston}]{yuan2024selfrewardinglanguagemodels}
Weizhe Yuan, Richard~Yuanzhe Pang, Kyunghyun Cho, Xian Li, Sainbayar
  Sukhbaatar, Jing Xu, and Jason Weston. 2024.
\newblock \href {https://arxiv.org/abs/2401.10020} {Self-rewarding language
  models}.
\newblock \emph{Preprint}, arXiv:2401.10020.

\bibitem[{Zhang et~al.(2024)Zhang, Dong, Li, Zhang, Sun, Wang, Li, Hu, Zhang,
  Wu, and Wang}]{zhang2024instructiontuninglargelanguage}
Shengyu Zhang, Linfeng Dong, Xiaoya Li, Sen Zhang, Xiaofei Sun, Shuhe Wang,
  Jiwei Li, Runyi Hu, Tianwei Zhang, Fei Wu, and Guoyin Wang. 2024.
\newblock \href {https://arxiv.org/abs/2308.10792} {Instruction tuning for
  large language models: A survey}.
\newblock \emph{Preprint}, arXiv:2308.10792.

\bibitem[{Zhang and Yang(2023)}]{zhang2023selfqaunsupervisedknowledgeguided}
Xuanyu Zhang and Qing Yang. 2023.
\newblock \href {https://arxiv.org/abs/2305.11952} {Self-qa: Unsupervised
  knowledge guided language model alignment}.
\newblock \emph{Preprint}, arXiv:2305.11952.

\bibitem[{Zheng et~al.(2024)Zheng, Zhang, Zhang, Ye, Luo, Feng, and
  Ma}]{zheng2024llamafactory}
Yaowei Zheng, Richong Zhang, Junhao Zhang, Yanhan Ye, Zheyan Luo, Zhangchi
  Feng, and Yongqiang Ma. 2024.
\newblock \href {http://arxiv.org/abs/2403.13372} {Llamafactory: Unified
  efficient fine-tuning of 100+ language models}.
\newblock In \emph{Proceedings of the 62nd Annual Meeting of the Association
  for Computational Linguistics (Volume 3: System Demonstrations)}, Bangkok,
  Thailand. Association for Computational Linguistics.

\bibitem[{Zhou et~al.(2023{\natexlab{a}})Zhou, Liu, Xu, Iyer, Sun, Mao, Ma,
  Efrat, Yu, Yu, Zhang, Ghosh, Lewis, Zettlemoyer, and
  Levy}]{zhou2023limaalignment}
Chunting Zhou, Pengfei Liu, Puxin Xu, Srini Iyer, Jiao Sun, Yuning Mao, Xuezhe
  Ma, Avia Efrat, Ping Yu, Lili Yu, Susan Zhang, Gargi Ghosh, Mike Lewis, Luke
  Zettlemoyer, and Omer Levy. 2023{\natexlab{a}}.
\newblock \href {https://arxiv.org/abs/2305.11206} {Lima: Less is more for
  alignment}.
\newblock \emph{Preprint}, arXiv:2305.11206.

\bibitem[{Zhou et~al.(2023{\natexlab{b}})Zhou, Lu, Mishra, Brahma, Basu, Luan,
  Zhou, and Hou}]{zhou2023instructionfollowingevaluationlargelanguage}
Jeffrey Zhou, Tianjian Lu, Swaroop Mishra, Siddhartha Brahma, Sujoy Basu,
  Yi~Luan, Denny Zhou, and Le~Hou. 2023{\natexlab{b}}.
\newblock \href {https://arxiv.org/abs/2311.07911} {Instruction-following
  evaluation for large language models}.
\newblock \emph{Preprint}, arXiv:2311.07911.

\bibitem[{Zhu et~al.(2018)Zhu, Lu, Zheng, Guo, Zhang, Wang, and
  Yu}]{zhu2018texygenbenchmarkingplatformtext}
Yaoming Zhu, Sidi Lu, Lei Zheng, Jiaxian Guo, Weinan Zhang, Jun Wang, and Yong
  Yu. 2018.
\newblock \href {https://arxiv.org/abs/1802.01886} {Texygen: A benchmarking
  platform for text generation models}.
\newblock \emph{Preprint}, arXiv:1802.01886.

\end{thebibliography}

\newpage
\appendix
\section{Datasets}
\label{sec:app_dataset}
\subsection{Dataset Details}
This section provides comprehensive details regarding the datasets utilized in this study.
\paragraph{SQuAD\citep{rajpurkar-etal-2016-squad}} This Dataset serves as a benchmark for extractive question answering, containing over 100,000 human-generated question-answer pairs anchored to Wikipedia passages. In our work, we utilize this dataset from the MRQA 2019 shared task\citep{fisch-etal-2019-mrqa}. Considering computational costs, we employ 2,500 randomly selected articles or contexts from the training set.
\paragraph{HotpotQA\citep{yang-etal-2018-hotpotqa}} The original purpose of proposing this dataset was to challenge models with multi-hop reasoning across 113k Wikipedia-based QA pairs requiring synthesis of information from multiple documents. In our work, we utilize this dataset from the MRQA 2019 shared task\citep{fisch-etal-2019-mrqa}. Considering computational costs, we employ 2,500 randomly selected articles or contexts from the training set.
\paragraph{FilmWiki\citep{wiki2023plus}} This dataset was initially constructed to investigate a phenomenon called the perplexity curse, with all texts sourced from Wikipedia. Our study employs its film-themed subset containing the complete collection of 2,385 unsupervised texts and their corresponding questions.

\subsection{Usage of Datasets}
During the synthesis process, we use the articles or contexts from the dataset (excluding the questions) and synthesize instructions through the \approach. In the evaluation of problem-solving capabilities, we assess the fine-tuned models using the corresponding questions in the dataset.
\section{Instruction Synthesis Details}
\label{sec:app_instruction_synthesis}
\label{sec:app_taskgen}

\subsection{Teacher Models and Sampling Hyperparameters}
\label{sec:app_teacher_models}
Taking into account both performance and API costs, and to eliminate potential biases in experimental results caused by specific teacher models (such as prompt hacking), we employ two models, GPT-4o mini(gpt-4o-mini-2024-07-18)\citep{openai2024gpt4technicalreport} and DeepSeek-V3\citep{deepseekai2025deepseekv3technicalreport}, to synthesize instructions from unsupervised text using the same methodology and sampling hyperparameters. Detailed hyperparameters are provided in \autoref{tab:app_taskgen_hype}, where, as described in Section \ref{sec:method}, the normal mode is utilized for the initial foveate synthesis, and the high-creativity mode is adopted for re-synthesis.

\begin{table}[h]
    \small
    \centering
    \renewcommand\arraystretch{1.1}
    \setlength{\tabcolsep}{6pt}
    \adjustbox{max width=\columnwidth}{
        \begin{tabular}{p{6.0cm}r}
            \toprule[1pt]
            \textbf{Hyperparameters} & \textbf{Values} \\
            \midrule
            \rowcolor{gray!10}
            \multicolumn{2}{c}{\textit{Normal Mode}} \\
            frequency\_penalty & 0.5 \\
            max\_completion\_tokens & None \\
            presence\_penalty & 0 \\
            temperature & 0.5 \\
            top\_p & 1.0 \\\midrule
            \rowcolor{gray!10}
            \multicolumn{2}{c}{\textit{High-creativity Mode}} \\
            frequency\_penalty & 0.5 \\
            max\_completion\_tokens & None \\
            presence\_penalty & 0 \\
            temperature & 1.2 \\
            top\_p & 1.0 \\
            \bottomrule[1pt]
        \end{tabular}
    }
    \caption{Hyperparameters of Synthetic Instruction Synthesis.}
    \label{tab:app_taskgen_hype}
\end{table}

\subsection{Adaptive Task Count Determination}
\label{sec:app_thresholding}
To adaptively determine the number of instructions to synthesize for each article, \approach employs a two-stage process: (1) counting contextually relevant words via semantic similarity thresholding, and (2) normalizing the resulting counts using a Box-Cox transformation. This ensures that articles with richer content receive proportionally more instructions while maintaining a balanced distribution across the dataset.

\paragraph{Context Thresholding.}
For each article $d_i$, we compute the text-level embedding $\mathbf{e}_{\text{text}} = \text{Embed}(d_i)$ and the word-level embeddings $\mathbf{e}_{w_j} = \text{Embed}(w_j)$ for each unique word $w_j$ in $d_i$. A word is considered contextually relevant if its cosine similarity with the full text exceeds the threshold $\tau$:
\begin{equation}
    c_i = \left|\left\{w_j : \text{CosSim}(\mathbf{e}_{\text{text}}, \mathbf{e}_{w_j}) \geq \tau \right\}\right|
\end{equation}
where $\tau = 0.16$ (\texttt{CONTEXT\_THRESHOLD}) by default.

\paragraph{Box-Cox Normalization.}
The raw counts $\{c_i\}$ are then normalized using the Box-Cox transformation to produce the final task counts. The procedure is detailed in Algorithm~\ref{alg:boxcox}.

\begin{algorithm}[h]
\caption{Adaptive Task Count via Box-Cox Normalization}
\label{alg:boxcox}
\small\setstretch{1.15}
\begin{algorithmic}[1]
\Require Article set $\mathcal{D} = \{d_1, \ldots, d_N\}$
\Statex \hspace{\algorithmicindent} Threshold $\tau = 0.16$, target mean $\mu = 8.0$
\Statex \hspace{\algorithmicindent} Scaling factor $\alpha = 1.0$
\Ensure Task counts $\{n_i\}$ for each article $d_i$
\For{each article $d_i \in \mathcal{D}$}
    \State $\mathbf{e}_{\text{text}} \gets \text{Embed}(d_i)$
    \State $W \gets \text{UniqueWords}(d_i)$
    \For{each word $w_j \in W$}
        \State $\mathbf{e}_{w_j} \gets \text{Embed}(w_j)$
        \State $s_j \gets \text{CosSim}(\mathbf{e}_{\text{text}}, \mathbf{e}_{w_j})$
    \EndFor
    \State $c_i \gets |\{w_j : s_j \geq \tau\}|$
\EndFor
\State \textit{// Ensure positivity for Box-Cox}
\If{$\min(\{c_i\}) \leq 0$}
    \State $c_i \gets c_i + |\min(\{c_i\})| + 1, \quad \forall\, i$
\EndIf
\State $\mathbf{c}^{\prime}, \hat{\lambda} \gets \text{BoxCox}(\{c_i\})$ \Comment{Optimal $\lambda$ via MLE}
\State $\bar{c}^{\prime} \gets \text{mean}(\mathbf{c}^{\prime}), \quad \sigma_{c^{\prime}} \gets \text{std}(\mathbf{c}^{\prime})$
\State $n_i \gets \text{round}\!\left(\mu + \alpha \cdot \frac{c^{\prime}_i - \bar{c}^{\prime}}{\sigma_{c^{\prime}}}\right), \quad \forall\, i$
\State \Return $\{n_i\}$
\end{algorithmic}
\end{algorithm}

\subsection{Scatter Keyword Grouping}
\label{sec:app_scatter_grouping}
The scatter-foveate level synthesizes instructions from keyword groups of varying sizes (1, 2, or 3 keywords per group). Given a target number of scatter-level instructions $n_{\text{scatter}}$ for an article, the number of groups at each size is determined by two configuration ratios, as described in Algorithm~\ref{alg:scatter_grouping}.

\begin{algorithm}[h]
\caption{Scatter Keyword Grouping}
\label{alg:scatter_grouping}
\small\setstretch{1.15}
\begin{algorithmic}[1]
\Require Target instruction count $n_{\text{scatter}}$, ratios $r_a = 2.50$, $r_b = 2.25$
\Ensure Keyword groups $\mathcal{G}$
\State \textit{// Compute group distribution}
\State $n_a \gets \lfloor n_{\text{scatter}} / r_a \rfloor$ \Comment{Single-keyword groups}
\State $n_b \gets \lfloor n_{\text{scatter}} / r_b \rfloor$ \Comment{Two-keyword groups}
\State $n_c \gets n_{\text{scatter}} - n_a - n_b$ \Comment{Three-keyword groups}
\State \textit{// Compute total keywords needed}
\State $n_{\text{total}} \gets n_a + 2 \cdot n_b + 3 \cdot n_c$
\State \textit{// Extract and rank keywords via LLM}
\State $K \gets \text{ExtractKeywords}(\text{article}, n_{\text{total}})$
\State $K_{\text{core}}, K_{\text{major}} \gets \text{RankImportance}(\text{article}, K)$
\State $K \gets \text{Shuffle}(K \cup K_{\text{core}} \cup K_{\text{major}})$
\State \textit{// Construct groups}
\State $\mathcal{G}_1 \gets \{\{k_i\} : i = 1, \ldots, n_a\}$
\State $\mathcal{G}_2 \gets \{\{k_i, k_{i+1}\} : \text{next } n_b \text{ pairs}\}$
\State $\mathcal{G}_3 \gets \{\{k_i, k_{i+1}, k_{i+2}\} : \text{next } n_c \text{ triples}\}$
\State \Return $\mathcal{G} = \mathcal{G}_1 \cup \mathcal{G}_2 \cup \mathcal{G}_3$
\end{algorithmic}
\end{algorithm}

The ratios $r_a$ (\texttt{SCATTER\_RATIO\_A} $= 2.50$) and $r_b$ (\texttt{SCATTER\_RATIO\_B} $= 2.25$) are designed to ensure a balanced distribution of keyword group sizes. Larger groups (3 keywords) produce more complex, multi-aspect instructions, while single-keyword groups yield focused, targeted questions. By including important (core and major) keywords into the shuffled pool, we ensure that critical concepts appear across multiple group sizes, further enhancing instruction diversity.

\subsection{Re-synthesis Module}
\label{sec:app_resynthesis}
The re-synthesis module handles cases where initial instruction generation fails, either due to invalid format or unanswerable content. The process follows a structured retry mechanism:

\paragraph{Validation Criteria.}
A generated instruction-answer pair is considered \textit{valid} if it satisfies all of the following:
\begin{enumerate}[itemsep=2pt, topsep=4pt, parsep=0pt]
    \item The output matches the expected ``\texttt{Question: ... Answer: ...}'' format (verified via regex parsing).
    \item The answer does not contain ``I don't know,'' indicating the model could not derive an answer from the background knowledge.
    \item The instruction and answer are both non-empty after stripping whitespace.
\end{enumerate}

\paragraph{Re-synthesis Flow.}
For each failed instruction, the re-synthesis module applies the retry procedure described in Algorithm~\ref{alg:resynthesis}.

\begin{algorithm}[h]
\caption{Re-synthesis Retry Procedure}
\label{alg:resynthesis}
\small\setstretch{1.15}
\begin{algorithmic}[1]
\Require Failed index set $\mathcal{F}$, successful pairs $\mathcal{S}$, article $d$, max retries $R = 5$
\Ensure Re-synthesized instruction set $\mathcal{I}_{\text{new}}$
\For{each failed index $i \in \mathcal{F}$}
    \State $f_i \gets \text{GetFeature}(i)$ \Comment{keyword / group / sentence}
    \For{$r = 1$ \textbf{to} $R$}
        \State $\mathcal{E} \gets \text{Shuffle}(\mathcal{S})$ \Comment{Few-shot examples}
        \State $(q, a) \gets \text{LLM}(d,\; f_i,\; \mathcal{E})$
        \If{$\text{IsValid}(q, a)$} \Comment{Format + answerable + non-empty}
            \State $\mathcal{I}_{\text{new}} \gets \mathcal{I}_{\text{new}} \cup \{(q, a)\}$
            \State \textbf{break}
        \EndIf
    \EndFor
\EndFor
\State \Return $\mathcal{I}_{\text{new}}$
\end{algorithmic}
\end{algorithm}

\subsection{Complete Hyperparameter Summary}
\label{sec:app_hyperparameters}
\autoref{tab:app_hyperparameters} provides a complete summary of all framework-level hyperparameters used in the \approach pipeline.

\begin{table*}[h]
    \small
    \centering
    \renewcommand\arraystretch{1.1}
    \setlength{\tabcolsep}{6pt}
    \adjustbox{max width=\textwidth}{
        \begin{tabular}{l@{\hskip 24pt}r@{\hskip 48pt}l}
            \toprule[1pt]
            \textbf{Parameter} & \textbf{Value} & \textbf{Description} \\
            \midrule
            \rowcolor{gray!10}
            \multicolumn{3}{c}{\textit{Thresholding \& Normalization}} \\
            \texttt{CONTEXT\_THRESHOLD} & 0.16 & Cosine similarity threshold for context relevance \\
            \texttt{DEFAULT\_TARGET\_MEAN} ($\mu$) & 8.0 & Target mean for Box-Cox normalization \\
            \texttt{DEFAULT\_ALPHA} ($\alpha$) & 1.0 & Scaling factor for Box-Cox transformation \\
            \midrule
            \rowcolor{gray!10}
            \multicolumn{3}{c}{\textit{Scatter Keyword Grouping}} \\
            \texttt{SCATTER\_RATIO\_A} & 2.50 & Ratio for single-keyword groups \\
            \texttt{SCATTER\_RATIO\_B} & 2.25 & Ratio for two-keyword groups \\
            \midrule
            \rowcolor{gray!10}
            \multicolumn{3}{c}{\textit{Processing Configuration}} \\
            \texttt{DEFAULT\_MAX\_RETRIES} & 5 & Maximum re-synthesis attempts \\
            \texttt{DEFAULT\_MAX\_WORKERS} & 96 & Parallel processing threads (articles) \\
            \texttt{EMBEDDING\_MAX\_WORKERS} & 10 & Parallel threads for embedding generation \\
            \bottomrule[1pt]
        \end{tabular}
    }
    \caption{Complete hyperparameter configuration of the \approach framework.}
    \label{tab:app_hyperparameters}
\end{table*}

\section{Prompt Templates -- Micro-foveate Level}
\label{sec:app_prompts_micro}

This section presents all prompt templates used in the \textbf{Micro-foveate} level of the \approach pipeline. The micro-level strategy extracts individual keywords from the article and synthesizes instructions based on those keywords.

\subsection{Keywords Extraction}

\begin{tcolorbox}[colback=blue!10!white,colframe=blue!50!black,title=Micro Keywords Extraction -- System Prompt]
\small
You are a professional text analysis expert. Your task is to extract keywords from the given article.\\
\\
Based on the following article, please extract \{num\_micro\} semantically different keywords in random order, with:\\
\\
Please ensure:\\
1. All keywords are based on the article content.\\
2. Each keyword is unique to maintain diversity.\\
3. Keywords can be single words or meaningful phrases.\\
4. Keywords should cover the main entities in the text and their related attributes or related entities.\\
5. Output strictly follows the format below without any other text:\\
1. keyword1\\
2. keyword2\\
...
\end{tcolorbox}

\begin{tcolorbox}[colback=blue!10!white,colframe=blue!50!black,title=Micro Keywords Extraction -- User Prompt]
\small
Article content:\\
\{content\}
\end{tcolorbox}

\subsection{Instruction Generation}

\begin{tcolorbox}[colback=blue!10!white,colframe=blue!50!black,title=Micro Instruction Generation -- System Prompt]
\small
You are an excellent questions/instructions generation expert. Your task is to generate \{num\_micro\} questions/instructions based on the given background knowledge and answers.\\
\\
Based on the following background knowledge and answers, please generate one question/instruction for each answer.\\
\\
Requirements:\\
1. Each generated question must be answerable by its corresponding answer\\
2. Questions/instructions must be based on the background knowledge content\\
3. Each question/instruction should focus on different aspects or information from the background knowledge\\
4. Output strictly follows JSON format as shown below without any other text:\\
\{json\_data\}
\end{tcolorbox}

\begin{tcolorbox}[colback=blue!10!white,colframe=blue!50!black,title=Micro Instruction Generation -- User Prompt]
\small
The background knowledge is:\\
\{content\}\\
\\
The answers are:\\
\{answers\}\\
\\
Please generate \{num\_micro\} questions/instructions based on the background knowledge and answers above.
\end{tcolorbox}

\subsection{Instruction Regeneration}

\begin{tcolorbox}[colback=blue!10!white,colframe=blue!50!black,title=Micro Instruction Regeneration -- System Prompt]
\small
You are an excellent questions/instructions generation expert. Your task is to generate one question/instruction for the target answer.\\
\\
Based on the following background knowledge and answers, please generate one question/instruction for the target answer.\\
\\
Requirements:\\
1. The generated question/instruction must be answerable by its corresponding answer.\\
2. The question/instruction must be based on the background knowledge content.\\
3. The question/instruction should be fully answerable using only the background knowledge.\\
4. Output the generated question/instruction only without any other text.
\end{tcolorbox}

\begin{tcolorbox}[colback=blue!10!white,colframe=blue!50!black,title=Micro Instruction Regeneration -- User Prompt]
\small
The background knowledge is:\\
\{content\}\\
\\
The target answer is:\\
\{answer\}
\end{tcolorbox}
\section{Prompt Templates -- Scatter-foveate Level}
\label{sec:app_prompts_scatter}

This section presents all prompt templates used in the \textbf{Scatter-foveate} level of the \approach pipeline. The scatter-level strategy extracts keyword groups of varying sizes (1, 2, or 3 keywords per group) and synthesizes instructions that incorporate all keywords from each group.

\subsection{Keywords Extraction}

\begin{tcolorbox}[colback=blue!10!white,colframe=blue!50!black,title=Scatter Keywords Extraction -- System Prompt]
\small
You are a professional text analysis expert. Your task is to extract keywords from the given article.\\
\\
Based on the following article, please extract \{num\_keywords\} semantically different keywords in random order, with:\\
\\
Please ensure:\\
1. All keywords are based on the article content.\\
2. Each keyword is unique to maintain diversity.\\
3. Keywords can be single words or meaningful phrases.\\
4. Keywords should cover the main entities in the text and their related attributes or related entities.\\
5. Output strictly follows the format below without any other text:\\
1. keyword1\\
2. keyword2\\
...
\end{tcolorbox}

\begin{tcolorbox}[colback=blue!10!white,colframe=blue!50!black,title=Scatter Keywords Extraction -- User Prompt]
\small
Article content:\\
\{content\}
\end{tcolorbox}

\subsection{Important Keywords Selection}

\begin{tcolorbox}[colback=blue!10!white,colframe=blue!50!black,title=Scatter Important Keywords Selection -- System Prompt]
\small
You are a professional text analysis expert. Your task is to identify the most important keywords from the given background knowledge.\\
\\
Please analyze the given keywords and select:\\
1. \{num\_core\} core keywords that are absolutely critical (primary keywords)\\
2. \{num\_major\} major keywords that are highly important but secondary (secondary keywords)\\
\\
For selecting the core keyword (select exactly \{num\_core\}):\\
- Must be the most essential concept that captures the central theme\\
- Should be impossible to understand the text without this keyword\\
- Removing this keyword would make the text lose its main focus\\
\\
For selecting major keywords (select exactly \{num\_major\}):\\
- Should be key supporting concepts that elaborate on the core keyword\\
- Must be directly relevant to explaining or contextualizing the main topic\\
- Should cover different important aspects of the content\\
\\
Please output strictly in this format without any other text:\\
Core keyword:\\
1. {[}first core keyword{]}\\
2. {[}second core keyword{]}\\
...\\
\\
Major keywords:\\
1. {[}first major keyword{]}\\
2. {[}second major keyword{]}\\
...
\end{tcolorbox}

\begin{tcolorbox}[colback=blue!10!white,colframe=blue!50!black,title=Scatter Important Keywords Selection -- User Prompt]
\small
Background knowledge:\\
\{content\}\\
\\
Keywords:\\
\{keywords\}
\end{tcolorbox}

\subsection{Instruction Generation}

\begin{tcolorbox}[colback=blue!10!white,colframe=blue!50!black,title=Scatter Instruction Generation -- System Prompt]
\small
You are an excellent questions/instructions generation expert. Your task is to generate \{num\_scatter\} questions/instructions based on the given background knowledge and keyword groups.\\
\\
Based on the following background knowledge and keyword groups, please generate one question/instruction for each keyword group that incorporates ALL keywords from that group.\\
\\
Requirements:\\
1. Each generated question/instruction must use ALL keywords from its corresponding group\\
2. Questions/instructions must be based on the background knowledge content\\
3. Each question/instruction should focus on different aspects or information from the background knowledge\\
4. Output strictly follows JSON format as shown below without any other text:\\
\{json\_data\}
\end{tcolorbox}

\begin{tcolorbox}[colback=blue!10!white,colframe=blue!50!black,title=Scatter Instruction Generation -- User Prompt]
\small
The background knowledge is:\\
\{content\}\\
\\
The keyword groups are:\\
\{keyword\_groups\}\\
\\
Please generate \{num\_scatter\} questions/instructions based on the background knowledge and keyword groups above.
\end{tcolorbox}

\subsection{Instruction Regeneration}

\begin{tcolorbox}[colback=blue!10!white,colframe=blue!50!black,title=Scatter Instruction Regeneration -- System Prompt]
\small
You are an excellent questions/instructions generation expert. Your task is to generate one question/instruction for the target keyword group that incorporates ALL keywords from that group.\\
\\
Based on the following background knowledge and keyword groups, please generate one question/instruction for the target keyword group that incorporates ALL keywords from that group.\\
\\
Requirements:\\
1. The generated question/instruction must use ALL keywords from its corresponding group.\\
2. The question/instruction must be based on the background knowledge content.\\
3. The question/instruction should be fully answerable using only the background knowledge.\\
4. Output the generated question/instruction only without any other text.
\end{tcolorbox}

\begin{tcolorbox}[colback=blue!10!white,colframe=blue!50!black,title=Scatter Instruction Regeneration -- User Prompt]
\small
The background knowledge is:\\
\{content\}\\
\\
The target keyword group to use:\\
\{keyword\_group\}
\end{tcolorbox}
\section{Prompt Templates -- Macro-foveate Level}
\label{sec:app_prompts_macro}

This section presents all prompt templates used in the \textbf{Macro-foveate} level of the \approach pipeline. The macro-level strategy extracts complete sentences as contextual features and synthesizes instructions based on those sentences.

\subsection{Sentence Extraction}

\begin{tcolorbox}[colback=blue!10!white,colframe=blue!50!black,title=Macro Sentence Extraction -- System Prompt]
\small
You are a professional text analysis expert. Your task is to extract sentences from the given article.\\
\\
Based on the following article, please extract \{num\_macro\} semantically different key sentences in random order, with:\\
\\
Please ensure:\\
1. All key sentences are based on the article content.\\
2. Each key sentence is unique to maintain diversity.\\
3. Key sentences should focus on identifying key details, focusing on transitions, quotations, comparisons, and rhetorical devices.\\
4. Key sentences should cover the main entities in the text and their related attributes or related entities.\\
5. Output strictly follows the format below without any other text:\\
1. key sentence1\\
2. key sentence2\\
...
\end{tcolorbox}

\begin{tcolorbox}[colback=blue!10!white,colframe=blue!50!black,title=Macro Sentence Extraction -- User Prompt]
\small
Article content:\\
\{content\}
\end{tcolorbox}

\subsection{Instruction Generation}

\begin{tcolorbox}[colback=blue!10!white,colframe=blue!50!black,title=Macro Instruction Generation -- System Prompt]
\small
You are an excellent questions/instructions generation expert. Your task is to generate \{num\_macro\} questions/instructions based on the given background knowledge and key sentences.\\
\\
Based on the following background knowledge and key sentences, please generate one question/instruction for each key sentence.\\
\\
Requirements:\\
1. Each generated question/instruction must use the corresponding key sentence.\\
2. Questions/instructions must be based on the background knowledge content.\\
3. Each question/instruction should focus on different aspects or information from the background knowledge.\\
4. Output strictly follows JSON format as shown below without any other text:\\
\{json\_data\}
\end{tcolorbox}

\begin{tcolorbox}[colback=blue!10!white,colframe=blue!50!black,title=Macro Instruction Generation -- User Prompt]
\small
The background knowledge is:\\
\{content\}\\
\\
The key sentences are:\\
\{sentences\}\\
\\
Please generate \{num\_macro\} questions/instructions based on the background knowledge and key sentences above.
\end{tcolorbox}

\subsection{Instruction Regeneration}

\begin{tcolorbox}[colback=blue!10!white,colframe=blue!50!black,title=Macro Instruction Regeneration -- System Prompt]
\small
You are an excellent questions/instructions generation expert. Your task is to generate one question/instruction for the target key sentence.\\
\\
Based on the following background knowledge and key sentences, please generate one question/instruction for the target key sentence.\\
\\
Requirements:\\
1. The generated question/instruction must use the corresponding key sentence.\\
2. The question/instruction must be based on the background knowledge content.\\
3. The question/instruction should be fully answerable using only the background knowledge.\\
4. Output the generated question/instruction only without any other text.
\end{tcolorbox}

\begin{tcolorbox}[colback=blue!10!white,colframe=blue!50!black,title=Macro Instruction Regeneration -- User Prompt]
\small
The background knowledge is:\\
\{content\}\\
\\
The target key sentence to use:\\
\{sentence\}
\end{tcolorbox}
\section{Prompt Templates -- Task Generation}
\label{sec:app_prompts_task}

This section presents all prompt templates used in the \textbf{Task Generation} module of the \approach pipeline. This module handles reading comprehension answer generation and answer improvement through rereading.

\subsection{Reading Comprehension (Answer Generation)}

\begin{tcolorbox}[colback=blue!10!white,colframe=blue!50!black,title=Reading Comprehension (Answer Generation) -- System Prompt]
\small
You are an excellent reading comprehension assistant designed to help users answer the question based on provided background knowledge. Please carefully read the background knowledge and accurately answer the question posed based on its content.\\
\\
- Do not alter key information from the original text.\\
- Avoid using phrases like ``based on the above article.''\\
- If the background knowledge lacks sufficient information to answer a question, respond with ``I don't know.''\\
- Ensure your answers are derived from the background knowledge, detailed, and accurate.\\
- Answer the question thoroughly.\\
\\
Please generate the answer following this format strictly:\\
Question: {[}question{]}\\
Answer: {[}detailed and accurate answer{]}
\end{tcolorbox}

\begin{tcolorbox}[colback=blue!10!white,colframe=blue!50!black,title=Reading Comprehension (Answer Generation) -- User Prompt]
\small
The background knowledge is:\\
\{unsupervised\_knowledge\_data\}\\
\\
Please answer the following question based on the content of the article above:\\
\{instruction\}
\end{tcolorbox}

\subsection{Answer Rereading (Answer Improvement)}

\begin{tcolorbox}[colback=blue!10!white,colframe=blue!50!black,title=Answer Rereading (Answer Improvement) -- System Prompt]
\small
You are a smart AI assistant. For a given question-answer pair, improve the answer by correcting errors, bolstering informativeness, aligning with the question, and providing comprehensive detail.\\
\\
- Do not alter key information from the original text.\\
- Avoid using phrases like ``based on the above article.''\\
- Ensure your answers are derived from the background knowledge, detailed, and accurate.\\
- Answer the question thoroughly.\\
\\
Please generate the rewritten answer following this format strictly:\\
Question: {[}question{]}\\
Answer: {[}detailed and accurate rewritten answer{]}
\end{tcolorbox}

\begin{tcolorbox}[colback=blue!10!white,colframe=blue!50!black,title=Answer Rereading (Answer Improvement) -- User Prompt]
\small
The background knowledge is:\\
\{unsupervised\_knowledge\_data\}\\
\\
The question is:\\
\{instruction\}\\
\\
The original answer is:\\
\{output\}\\
\\
Please reread the background knowledge and generate a more detailed and accurate rewritten answer based on the content of the article above.
\end{tcolorbox}
\section{Diversity Experiment}
\label{sec:app_diversity}
Mathematically, we define the diversity metrics as follows ('D' denotes 'Diversity', 'SB' denotes 'SelfBLEU', and 'EB' denotes 'Embedding'):
\begin{align}
\label{eq:div_selfbleu}
    \text{D}_{\text{SB}} &= 1 - \dfrac{1}{\lvert \mathcal{X}_{\tau} \rvert}
    \sum_{x_i \in \mathcal{X}_{\tau}}\sum^{5}_{n = 2} \text{SelfBLEU}_{\mathcal{X}_\tau}(x_i, n) \\
\label{eq:div_embd}
    \text{D}_{\text{EB}} &= 1 - \dfrac{2}{\lvert \mathcal{X}_{\tau} \rvert (\lvert \mathcal{X}_{\tau} \rvert - 1)} \sum_{i < j} \dfrac{\phi(x_i) \cdot \phi(x_j)}{\lVert \phi(x_i) \rVert \lVert \phi(x_j) \rVert}
\end{align}
For computing sentence embeddings, we employ sentence-transformers/paraphrase-multilingual-mpnet-base-v2\citep{reimers-2019-sentence-bert}, which maps sentences to a 768-dimensional dense vector space. Compared to other sentence embedding transformation models, it offers a balanced mapping dimension and accuracy, enabling the computation of cosine similarity matrices at an acceptable computational cost while obtaining relatively accurate diversity evaluation metrics.
\section{Difficulty Head-to-Head Experiment}
\label{sec:app_difficulty}
We prompt \sotallm(gpt-4o-2024-08-06)\citep{openai2024gpt4technicalreport} to determine which method-synthesized instruction set is more challenging. The prompts are shown below.

\begin{tcolorbox}[colback=blue!10!white,colframe=blue!50!black,title=Difficulty Head-to-Head Experiment\\
System Prompt]
\small
Please compare two sets of instructions/questions for a given content to determine which set is more difficult for a language model to generate coherent and accurate responses. A set of instructions is considered more difficult if it requires:\\
1. More complex reasoning or multi-step analysis\\
2. Integration of multiple concepts or contexts\\
3. Explanation of nuanced relationships or processes\\
4. Handling of ambiguous or less-documented information\\
5. Synthesis of information from diverse sources or time periods\\
\\
Please respond with '1' if the first set of instructions is more difficult, or '2' if the second set of instructions is more difficult.\\
Please respond with the number only, no other text or characters.
\end{tcolorbox}

\begin{tcolorbox}[colback=blue!10!white,colframe=blue!50!black,title=Difficulty Head-to-Head Experiment\\
User Prompt]
\small
Content: \{unsupervised text\}\\
Instruction Set 1: \{instruction set a\}\\
Instruction Set 2: \{instruction set b\}
\end{tcolorbox}

The hyperparameters set during sampling are listed in \autoref{tab:app_difficulty_hype}.

\begin{table}[h]
    \small
    \centering
    \renewcommand\arraystretch{1.1}
    \setlength{\tabcolsep}{6pt}
    \adjustbox{max width=\columnwidth}{
        \begin{tabular}{p{6.0cm}r}
            \toprule[1pt]
            \textbf{Hyperparameters} & \textbf{Values} \\
            \midrule
            frequency\_penalty & 0.0 \\
            max\_completion\_tokens & 5 \\
            presence\_penalty & 0 \\
            temperature & 0.5 \\
            top\_p & 1.0 \\
            \bottomrule[1pt]
        \end{tabular}
    }
    \caption{Hyperparameters of Difficulty Head-to-Head Experiment.}
    \label{tab:app_difficulty_hype}
\end{table}

\section{Downstream Experiment}
\label{sec:app_downstream}
\subsection{Software and Hardware Details}
The implementation leverages the LLaMA-Factory framework \citep{zheng2024llamafactory} with computational optimizations from FlashAttention2 \citep{dao2022flashattention, dao2023flashattention2} and Unsloth libraries \citep{unsloth}. For training, we employ NVIDIA A100, A800, and H800 GPUs based on availability within our computational cluster. The entire work required approximately 110 GPU hours.

\subsection{Hyperparameters}
The hyperparameters and technical configurations for the instruction tuning process are documented in \autoref{tab:app_downstream_hype}.

\begin{table}[h]
    \small
    \centering
    \renewcommand\arraystretch{1.1}
    \setlength{\tabcolsep}{6pt}
    \adjustbox{max width=\columnwidth}{
        \begin{tabular}{p{6.0cm}r}
            \toprule[1pt]
            \textbf{Hyperparameters} & \textbf{Values} \\
            \midrule
            cutoff\_len & 2048 \\
            learning\_rate & 0.0001 \\
            num\_train\_epochs & 5.0 \\
            effective\_batch\_size & 16 \\
            lr\_scheduler\_type & cosine \\
            max\_grad\_norm & 1.0 \\
            warmup\_steps & 0 \\
            optim & adamw\_torch \\
            quantization\_bit & 4 \\
            quantization\_method & bitsandbytes \\
            lora\_rank & 8 \\
            lora\_alpha & 16 \\
            lora\_dropout & 0 \\
            \bottomrule[1pt]
        \end{tabular}
    }
    \caption{Hyperparameters of Instruction Tuning with Q-LoRA Quantization.}
    \label{tab:app_downstream_hype}
\end{table}

\subsection{LLM Accuracy Evaluation}
\begin{tcolorbox}[colback=blue!10!white,colframe=blue!50!black,title=LLM Accuracy Evaluation\\
System Prompt]
\small
You are a fair judge. Your task is to determine if the generated answer correctly answers the question, even if it contains additional explanations.
Rules:\\
1. The generated answer is correct if it contains the key information from the ground truth\\
2. Additional explanations or context in the generated answer should not make it incorrect\\
3. Only respond with 'Correct' or 'Incorrect'
\end{tcolorbox}

\begin{tcolorbox}[colback=blue!10!white,colframe=blue!50!black,title=LLM Accuracy Evaluation\\
User Prompt]
\small
Compare the following answers:\\
Question: \{question\}\\
Ground Truth Answer: \{ground\_truth\}\\
Generated Answer: \{generated\}\\
\\
Is the generated answer correct, regardless of any additional explanation? Respond only with 'Correct' or 'Incorrect'.
\end{tcolorbox}
\section{Answer Regeneration Evaluation}
\label{sec:app_regeneration}
This section provides detailed documentation of the evaluation methodology used in the ablation study for answer regeneration analysis (Section \ref{exp:ablation}). To assess the quality of answers generated with and without the regeneration mechanism, we employed GPT-4o (gpt-4o-2024-08-06) to evaluate fluency and completeness on a three-point scale.

\subsection{Evaluation Prompt}
The following system prompt was used to evaluate question-answer pairs:

\begin{tcolorbox}[colback=blue!10!white,colframe=blue!50!black,title=Answer Regeneration Evaluation\\
System Prompt]
\small
Please evaluate the given question and answer pair based on two criteria:\\
\\
1. Fluency: How well does the answer flow and connect with the question?\\
\phantom{1.} - High: The answer naturally follows from the question\\
\phantom{1.} - Medium: The connection is somewhat clear but could be improved\\
\phantom{1.} - Low: The answer feels disconnected from the question\\
\\
2. Completeness: How thoroughly does the answer address the question?\\
\phantom{2.} - High: The answer fully addresses all aspects of the question\\
\phantom{2.} - Medium: The answer covers most aspects but misses some points\\
\phantom{2.} - Low: The answer only partially addresses the question\\
\\
Please respond with a JSON object in the following format:\\
\{\\
\phantom{....}"fluency": "high|medium|low",\\
\phantom{....}"completeness": "high|medium|low"\\
\}\\
\\
Do not include any other text or explanation.
\end{tcolorbox}

\subsection{Evaluation Hyperparameters}
The hyperparameters used for the answer regeneration evaluation are provided in \autoref{tab:app_regeneration_hype}.

\begin{table}[h]
    \small
    \centering
    \renewcommand\arraystretch{1.1}
    \setlength{\tabcolsep}{6pt}
    \adjustbox{max width=\columnwidth}{
        \begin{tabular}{p{6.0cm}r}
            \toprule[1pt]
            \textbf{Hyperparameters} & \textbf{Values} \\
            \midrule
            frequency\_penalty & 0.0 \\
            max\_completion\_tokens & 50 \\
            presence\_penalty & 0 \\
            temperature & 0.2 \\
            top\_p & 1.0 \\
            \bottomrule[1pt]
        \end{tabular}
    }
    \caption{Hyperparameters for Answer Regeneration Evaluation.}
    \label{tab:app_regeneration_hype}
\end{table}

\section{Use of AI Assistants}
In this study, we utilized AI-powered tools, including ChatGPT, to improve the linguistic quality of the manuscript through spell-checking and minor grammatical corrections. Additionally, the codebase was developed using Cursor to enhance coding efficiency and accuracy.              

\end{document}